\documentclass{article}

\PassOptionsToPackage{numbers, compress}{natbib}

\usepackage[preprint]{neurips_2026}

\usepackage[utf8]{inputenc} 
\usepackage[T1]{fontenc}    
\usepackage{hyperref}       
\usepackage{url}            
\usepackage{booktabs}       
\usepackage{amsfonts}       
\usepackage{amsmath}        
\usepackage{amssymb}        
\usepackage{nicefrac}      
\usepackage{microtype}      
\usepackage{xcolor}         
\usepackage{graphicx}
\usepackage{algorithm}
\usepackage{algpseudocode}
\usepackage[most]{tcolorbox}
\usepackage{lipsum}
\usepackage{kotex}
\usepackage{soul}
\usepackage{subcaption}
\usepackage{tabularx}
\usepackage{wrapfig}
\usepackage{xcolor}
\usepackage[most]{tcolorbox}

\usepackage{wrapfig}
\usepackage{xcolor}
\usepackage{booktabs}        
\usepackage[most]{tcolorbox}

\definecolor{incontextHL}{RGB}{181,212,244}
\definecolor{structuralHL}{RGB}{211,209,199}
\definecolor{storedHL}{RGB}{250,199,117}

\definecolor{incontextDark}{RGB}{24,95,165}
\definecolor{structuralDark}{RGB}{95,94,90}
\definecolor{storedDark}{RGB}{186,117,23}

\newcommand{\hlinc}[1]{\colorbox{incontextHL}{\strut #1}}
\newcommand{\hlstr}[1]{\colorbox{structuralHL}{\strut #1}}
\newcommand{\hlstk}[1]{\colorbox{storedHL}{\strut #1}}

\definecolor{deltaPlus}{RGB}{59,109,17}    
\definecolor{deltaMinus}{RGB}{163,45,45}   

\newcommand{\dpos}[1]{\,{\scriptsize\textcolor{deltaPlus}{\textbf{#1}}}}
\newcommand{\dneg}[1]{\,{\scriptsize\textcolor{deltaMinus}{\textbf{#1}}}}

\usepackage{booktabs}
\usepackage{tabularx}
\usepackage{xcolor}
\usepackage{makecell}

\usepackage{multirow}
\usepackage[table]{xcolor}

\definecolor{ctxblue}{RGB}{224,238,255}
\definecolor{ctxtext}{RGB}{30,80,140}
\definecolor{newgold}{RGB}{255,241,199}
\definecolor{newtext}{RGB}{145,95,20}
\definecolor{maskgray}{RGB}{238,238,238}
\definecolor{masktext}{RGB}{110,110,110}

\definecolor{rolloutfixedgray}{RGB}{242,242,242}

\newcommand{\rollfixed}[1]{%
  \begingroup
  \setlength{\fboxsep}{1.2pt}%
  \colorbox{rolloutfixedgray}{\strut\texttt{#1}}%
  \endgroup
}

\newcolumntype{L}{>{\raggedright\arraybackslash}X}

\title{Machine Unlearning for\\Masked Diffusion Language Models}

\author{%
  Georu Lee$^{1}$,
  Seungwon Jeong$^{1}$,
  Hoki Kim$^{2}$,
  Jinseong Park$^{3}$,
  Woojin Lee$^{1}$ \\
  $^{1}$Dongguk University-Seoul,
  $^{2}$Chung-Ang University,
  $^{3}$Korea Institute for Advanced Study \\
  \texttt{\{dlrjfn1,youai058,wj926\}@dgu.ac.kr} \\
  \texttt{hokikim@cau.ac.kr},\,\texttt{jinseong@kias.re.kr}
}

\begin{document}

\maketitle

\begin{abstract}
Recent masked diffusion language models (MDLMs), such as LLaDA and Dream, have achieved performance comparable to autoregressive large language models. Unlike autoregressive models, which generate text sequentially, MDLMs generate text by iteratively denoising masked positions in parallel. During fine-tuning, MDLMs learn to recover responses from masked response states conditioned on a prompt, thereby shifting their predictions from a prompt-masked unconditional distribution toward a prompt-conditional distribution. Despite this distinct generative and fine-tuning mechanism, machine unlearning for MDLMs remains largely unexplored.
In this paper, we propose \textbf{Masked Diffusion Unlearning (MDU)}, the first unlearning framework for MDLMs, by revisiting the process of learning specific knowledge in terms of diffusion. Specifically, MDU minimizes a forward KL divergence from the prompt-conditional prediction to a prompt-masked unconditional anchor at every masked response position, with a temperature scaling parameter to control the privacy-utility trade-off. Our empirical results on standard benchmarks and MDLM backbones show that MDU achieves high unlearning performance compared to existing LLM unlearning methods. Code is available at \url{https://github.com/leegeoru/MDU}.
\end{abstract}

\section{Introduction}
\label{sec:intro}

Autoregressive Large Language Models (LLMs) have become the dominant paradigm for modern text generation, achieving strong performance across a wide range of natural language tasks. This left-to-right factorization requires tokens to be generated one after another, with each prediction depending only on the preceding context. To move beyond this fixed sequential generation process, masked diffusion language models (MDLMs) have recently emerged as an alternative to the autoregressive paradigm~\cite{sahoo2024simple}. In this paradigm, generation proceeds through masked denoising states: the model starts from a fully masked sequence and repeatedly predicts masked positions in parallel rather than following a fixed token order. Recent MDLMs such as LLaDA~\cite{nie2025large} and Dream~\cite{ye2025dream} have scaled to billions of parameters, achieving performance competitive with autoregressive LLMs.

As LLMs, including emerging MDLMs, become increasingly capable and deployment-relevant, privacy, copyright, and regulatory concerns make it necessary to selectively remove specific data or knowledge after training. To address these concerns, machine unlearning has emerged as a principled framework for removing the influence of specific training data without retraining from scratch~\cite{cao2015towards,golatkar2020eternal}. Since autoregressive LLMs have dominated modern text generation, most machine unlearning research for language models has focused on this setting~\cite{yao2024large,zhang2024negative}. Meanwhile, recent work has begun to study safety in MDLMs, analyzing jailbreaking vulnerabilities and developing alignment methods that account for their parallel generation and bidirectional dependencies~\cite{zhang2025jailbreaking,li2025diffuguard,jeung2025a2d}. However, machine unlearning for MDLMs remains largely unexplored, leaving open how to design unlearning objectives for masked denoising states rather than sequential next-token prediction.


In this work, we formulate unlearning in MDLMs as reversing the trajectory-level shift induced by diffusion-based fine-tuning. During fine-tuning, an MDLM trained on a prompt--response pair $(x, y)$ learns to recover the response $y$ from masked response states $y_t$ conditioned on the prompt $x$. This process shifts the denoising trajectory at each masked response position from a prompt-masked unconditional distribution $p_\theta(\cdot \mid m, y_t)$ toward a prompt-conditional distribution $p_\theta(\cdot \mid x, y_t)$. From this perspective, the process of learning specific knowledge corresponds to a divergence of the conditional denoising trajectory from its prompt-masked unconditional anchor.


Therefore, we propose \textbf{Masked Diffusion Unlearning (MDU)}, the first unlearning objective designed for MDLMs.
For each forget pair, MDU minimizes a forward Kullback-Leibler (KL) divergence from the prompt-conditional prediction to a temperature-scaled prompt-masked anchor at masked response positions sampled along the denoising process, where the temperature controls the privacy-utility trade-off.
Token-level and convergence analyses show that MDU primarily affects the prompt-conditional knowledge targeted for removal, while preserving the broader behavior and converging toward the intended unlearning target.
On TOFU~\citep{maini2024tofu} and RWKU~\citep{cao2024rwku} with LLaDA-8B~\citep{nie2025large} and Dream-7B~\citep{ye2025dream}, the proposed framework achieves strong unlearning performance compared with existing LLM unlearning methods, highlighting the potential of MDLM-adaptive unlearning.

\section{Related Work}
\paragraph{Masked Diffusion Language Model.}
The idea of adapting diffusion processes to discrete domains stems from discrete-state generative diffusion~\citep{sohl2015deep, hoogeboom2021argmax}. These early approaches were later unified into structured frameworks~\citep{austin2021structured},
connected to continuous-time formulations~\citep{campbell2022continuous},
and extended through likelihood-ratio estimation~\citep{lou2023discrete}. Mask-based processes further led to clean masked-token denoising objectives~\citep{sahoo2024simple, shi2024simplified}. In MDLMs, generation proceeds from a fully masked sequence through partially masked denoising states, where the model predicts vocabulary distributions for masked positions rather than predicting a single next-token with a left-to-right prefix. This paradigm relaxes the fixed generation order of autoregressive LLMs and enables iterative refinement over masked positions. Building on this foundation, MDLMs have scaled to billions of parameters and achieved performance competitive with autoregressive baselines~\citep{nie2025large, ye2025dream}. Recent work on decoding~\citep{kim2025train}, alignment, and safety~\citep{zhang2025jailbreaking, li2025diffuguard, jeung2025a2d} has shown that MDLMs require methods that account for their parallel and bidirectional denoising structure. However, unlike decoding and safety, machine unlearning has not yet been formulated for MDLMs. 

\paragraph{Machine Unlearning.}
Machine unlearning \cite{cao2015towards, golatkar2020eternal} fundamentally aims to eliminate the influence of forget data $\mathcal{D}_f$ while preserving knowledge learned from retain data $\mathcal{D}_r$. Since retraining models from scratch (\textit{exact unlearning}) is computationally prohibitive for large-scale deep learning models \cite{zemel2013learning, cha2024learning}, various \textit{approximate unlearning} methods have been developed to efficiently achieve this goal \cite{golatkar2020eternal}.
Building upon these foundational concepts, machine unlearning has been actively explored and adapted across diverse domains. In LLMs, recent approaches focus on weak unlearning \cite{kurmanji2023towards} to modify model behavior without full retraining, typically by aligning or constraining internal representations to enforce forgetting \cite{fan2023salun, li2024wmdp}. Similarly, in the visual generation domain, unlearning techniques for latent diffusion models have been studied to erase harmful concepts~\citep{gandikota2023erasing}. These studies continuously evolve to address complex challenges such as robustness-utility trade-offs and the unintended recovery of erased concepts via textual inversion~\citep{zhang2024defensive, srivatsan2025stereo}.

\section{Background}
\label{sec:prelim-dlm}
Masked diffusion language models are mask predictors $p_\theta$, networks trained to predict masked tokens from a partially masked input. Following LLaDA~\citep{nie2025large}, we sample a clean sequence $x_0 \sim \mathcal{D}_{\mathrm{pre}}$ and a mask ratio $t \sim \mathcal{U}[0, 1]$, and mask each token independently with probability $t$ under the forward masking distribution $q(\cdot \mid x_0, t)$, producing a partially masked sequence $x_t$. Let $\mathcal{M}_t = \{i : x_t^i = \texttt{MASK}\}$ denote the set of masked positions. The model outputs probability distributions $p_\theta(\cdot \mid x_t)$ over the vocabulary at every position in $\mathcal{M}_t$ in parallel.
Training MDLMs \cite{nie2025large} mainly consists of pre-training and fine-tuning stages. In pre-training, based on the masked token, we train the model by minimizing the masked-token loss as follows:
{{\small\begin{equation}
\mathcal{L}_{\mathrm{pre}}(\theta;\mathcal{D}_{\mathrm{pre}}) = -\mathbb{E}_{x_0,\, t,\, x_t}\left[\frac{1}{t}\sum_{i\in \mathcal{M}_t}\log p_\theta(x_0^i \mid x_t)\right] = \mathbb{E}_{x_0,\, t,\, x_t}\left[\frac{1}{t}\sum_{i\in \mathcal{M}_t} \mathrm{KL}\Big( q(x_0^i) \,\big\|\, p_\theta(\cdot \mid x_t) \Big)\right],
\label{eq:pre}
\end{equation}}
where $x_0 \sim \mathcal{D}_{\mathrm{pre}}$ is a clean sequence sampled from the pre-training dataset, $t \sim \mathcal{U}[0,1]$ is the uniformly sampled mask ratio, and $x_t \sim q(\cdot \mid x_0, t)$ is the partially masked sequence.

In fine-tuning stages, the model is trained on the conditional generation of a response $y=(y_1,\ldots,y_n)$ given a prompt $x$, where $(x, y)$ is sampled from a fine-tuning dataset $\mathcal{D}$. We sample a mask ratio $t \sim \mathcal{U}[0, 1]$ and independently mask each response token with probability $t$ under $q(\cdot \mid y, t)$, producing a partially masked state $y_t$ while the prompt remains visible. As before, let $\mathcal{M}_t = \{i : y_t^i = \texttt{MASK}\}$ denote the masked positions. At $t=1$, the response is fully masked, and as $t \to 0^+$, $y_t$ approaches the clean response $y$. The model outputs probability distributions $p_\theta(\cdot \mid x, y_t)$ over the vocabulary at every position in $\mathcal{M}_t$. Fine-tuning loss is similar to pre-training, but based on the response tokens:
{\small\begin{equation}
\mathcal{L}_{\mathrm{sft}}(\theta;\mathcal{D}) = -\mathbb{E}_{(x,y),\, t,\, y_t}\left[\frac{1}{t}\sum_{i\in \mathcal{M}_t}\log p_\theta(y_i\mid x,y_t)\right] =\mathbb{E}_{(x,y),\, t,\, y_t}\left[\frac{1}{t}\sum_{i\in \mathcal{M}_t} \mathrm{KL}\Big( q(y_i) \,\big\|\, p_\theta(\cdot \mid x, y_t) \Big)\right],
\label{eq:dlm}
\end{equation}}
where $(x, y) \sim \mathcal{D}$ is sampled from the fine-tuning dataset, and $y_t \sim q(\cdot \mid y, t)$ is the partially masked response. We refer to the pair $(x, y_t)$, the prompt together with a partially masked response, as the \emph{masked denoising state} at noise level $t$. At inference, generation starts from the fully masked state $y_1 = [\texttt{MASK}]^n$, and the same model is applied at every step to the current denoising state $(x, y_t)$, unmasking positions until an unmasked response is produced.

\section{Masked Diffusion Unlearning} \label{sec:mdu}

In this section, we revisit the process of learning specific knowledge in the context of MDLMs. Given a dataset $\mathcal{D}$ of prompt-response pairs $(x, y)$, a model $p_\theta$ learns to encode specific knowledge through its conditional denoising distribution. Unlike autoregressive LLMs, diffusion models acquire such information by shifting their denoising trajectory away from a reference trajectory. Here, the reference trajectory corresponds to $p_\theta(\cdot \mid \emptyset)$ in the domain of general diffusion, often referred to as the null or unconditional trajectory, which represents the marginal distribution of the full dataset \cite{mokady2023null, bouvier2025ddat, stoica2025contrastive}. In continuous diffusion models, this reference trajectory is commonly understood to preserve an average statistical representation or structural blueprint of the data, capturing the global density of the data manifold rather than instance-specific details \cite{mokady2023null, rombach2022high}. This property is widely exploited for optimization and sampling, most notably in classifier-free guidance \cite{ho2022classifier}.

In the language domain, this reference trajectory can be formalized as $p_\theta(\cdot \mid m, y_t)$, where $m$ denotes a null prompt or mask. We view the process of learning specific knowledge, such as facts or entities, not merely as pattern matching, but as an optimization process that induces a new trajectory $p_\theta(\cdot \mid x, y_t)$ from the reference trajectory. Consequently, to conceptualize the process of learning prompt-response pairs $(x, y_t)$, we formalize it as a distributional shift from mask-response pairs $(m,y_t)$ as follows:
\begin{equation}
p_\theta(\cdot \mid m, y_t) \xrightarrow{\text{div.}} p_\theta(\cdot \mid x, y_t),
\end{equation}
where $\xrightarrow{\mathrm{div.}}$ denotes the divergence (or separation) from the null trajectory. As in Eq.~\eqref{eq:dlm}, note that recent work leveraged this conceptualization for supervised fine-tuning in MDLMs \cite{nie2025large, nie2025scaling}.

Machine unlearning aims to remove specific knowledge, such as sensitive concepts or copyrighted entities, from the model parameters. For a forget set $\mathcal{D}_f$ containing pairs $(x_f, y_f)$, successful unlearning requires that $p_\theta(\cdot \mid x_f, y_t)$ no longer assigns high probability to the memorized response $y_f$. Building on this, we formalize unlearning in MDLMs as driving the conditional prediction back toward the unconditional state. Rather than introducing a separate forgetting distribution, we can optimize the model to collapse the specific conditional trajectory onto the marginal reference trajectory:
\begin{equation}
\label{eq:mapsto_unlearn}
p_\theta(\cdot \mid x, y_t) \xrightarrow{\text{conv.}} p_\theta(\cdot \mid m, y_t),
\end{equation}
where $\xrightarrow{\mathrm{conv.}}$ denotes the convergence back (or merge) to the null trajectory.
By enforcing this objective, the model is forced to treat the prompt $x_f$ as uninformative, thereby reverting to its general linguistic prior and effectively erasing the specific mutual information previously associated with $x_f$.

\begin{figure}[t]
    \centering
    \includegraphics[width=\linewidth]{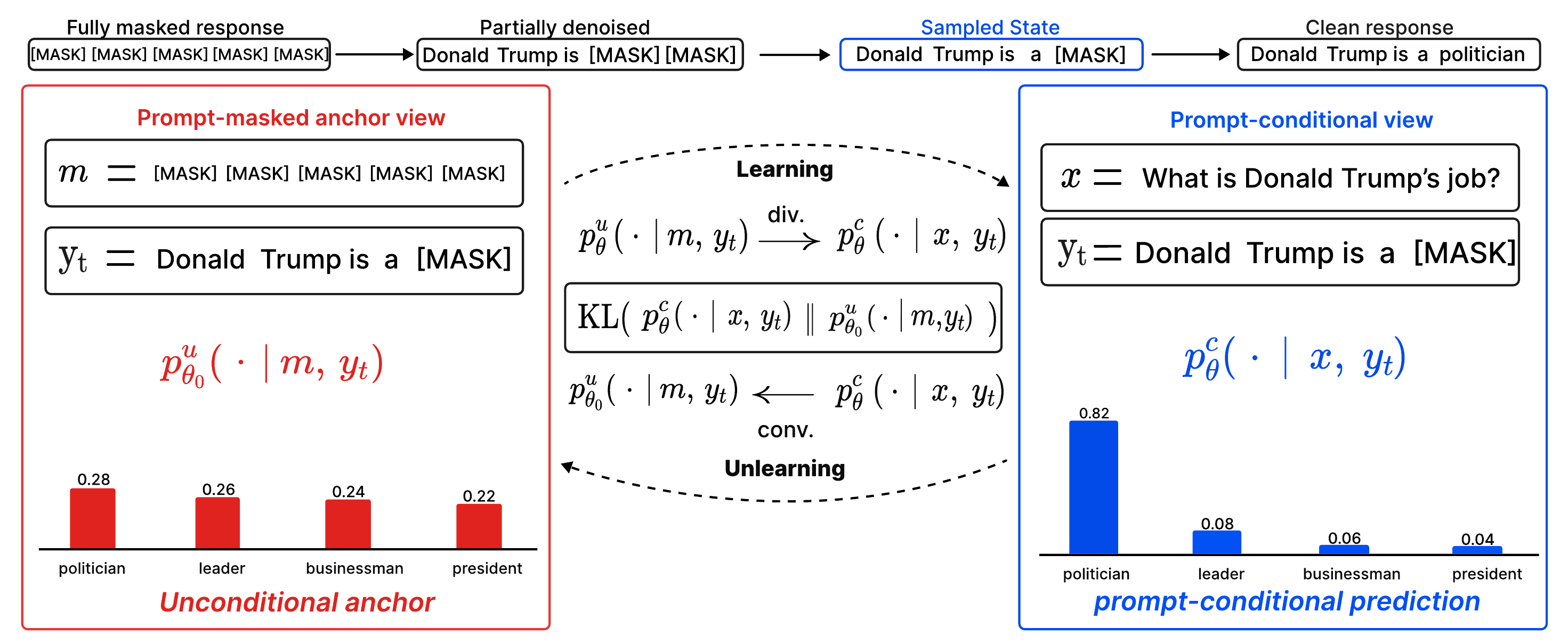}
    \caption{Overview of the proposed Masked Diffusion Unlearning (MDU).}
    \label{fig:mdu}
\end{figure}

Thus, we introduce \textbf{Masked Diffusion Unlearning (MDU)}, which minimizes the KL divergence between the conditional \textbf{answer} and the unconditional \textbf{anchor} at every masked response position as
\begin{equation}
\label{eq:loss}
\mathcal{L}_{forget}(\theta) = \mathbb{E}_{(x,y) \sim \mathcal{D}_f,\, t,\, y_t} \left[ \frac{1}{|\mathcal{M}_t|} \sum_{i \in \mathcal{M}_t} \mathrm{KL}\!\left( p^c_\theta(\cdot \mid x, y_t) \,\Big\|\, \frac{1}{Z_i}\, p^u_{\theta_0}(\cdot \mid m, y_t)^{\tau} \right) \right].
\end{equation}
Here, $\theta_0$ denotes the model parameters at the start of unlearning, where $Z_i = \sum_v p^u_{\theta_0}(v \mid m, y_t)^{\tau}$ is the normalization constant, and $\tau \in [0, 1]$ denotes the temperature scale of an anchor. 
In practice, we add a weighted reconstruction loss $\mathcal{L}_{retain}$ on a retain set $\mathcal{D}_r$ to preserve unrelated knowledge.  At each training step, we sample a random mask ratio $t \sim \mathcal{U}[0, 1]$, draw $y_t \sim q(\cdot \mid y, t)$, and perform forward passes, one with the prompt $x$ and one with its mask sequence $m$, to compute the per-position KL in \eqref{eq:loss}. We provide the detailed unlearning scheme in Algorithm~\ref{alg:mdu} (Appendix~\ref{app:mdu-algo}).

Beyond a straightforward KL formulation, the objective in Eq.~\eqref{eq:loss}
unifies existing unlearning objectives. First, we can interpret the discrepancy
between the conditional distribution $p_\theta(y_i \mid x, y_t)$ and the
unlearning target distribution $p_{\theta_0}(y_i \mid m, y_t)^{\tau}$ in
Eq.~\eqref{eq:loss} as a guidance toward a tilted distribution:
\begin{equation}\label{eq:cfg_dlm}
\tilde{p}_\theta(y_i \mid x, y_t)
\propto
p_\theta(y_i \mid x, y_t)^{1-\tau}
p_{\theta_0}(y_i \mid m, y_t)^{\tau}.
\end{equation}
This is related to unsupervised classifier-free guidance (CFG)
\cite{ho2022classifier} in discrete diffusion \cite{nie2025scaling},
$\tilde{p}_\theta(y_i \mid x, y_t) \propto
p_\theta(y_i \mid x, y_t)^{1+w}
p_\theta(y_i \mid m, y_t)^{-w}$, where $w>0$ denotes the guidance strength.
Unlike standard CFG, which combines conditional and unconditional predictions to
amplify the conditional signal, Eq.~\eqref{eq:cfg_dlm} interpolates between the
conditional distribution and the fixed prompt-masked anchor.
In text-to-image unlearning tasks, removing the conditional guidance has already
been established in Erased Stable Diffusion (ESD)~\cite{gandikota2023erasing},
and several papers~\cite{wen2024detecting,jeon2025understanding,jain2025classifier}
also argue that CFG-related signals provide useful clues for memorization.

Second, controlling our temperature $\tau$ gives the flexibility to select the
desired target distribution for unlearning. When $\tau=1$, as in
Eq.~\eqref{eq:mapsto_unlearn}, the target anchor distribution becomes
$p_{\theta_0}(y_i \mid m, y_t)$. However, it sometimes under-forgets information that needs to be forgotten. When $\tau = 0$, the target anchor distribution becomes a uniform distribution, which is exactly the same as the Maximum Entropy (ME) loss \cite{yuan2024closer}. This forces high entropy on forget samples to unlearn, but might suffer from over-forgetting. Thus, the proposed MDU can control the privacy--utility trade-off with a single hyperparameter $\tau$.

\section{Analysis of Denoising Behavior and Anchors}
\label{sec:behavior}
\label{sec:mldm}
\label{sec:Rollouts}

We now experimentally show that the prompt-masked anchor defined in Section~\ref{sec:mdu} provides a meaningful reference for isolating the prompt-conditional signal in MDLM unlearning.

\paragraph{Responses from Partially Denoised States}
For the prompt-masked anchor to serve as a meaningful reference for the forget objective, we should remove the prompt information that supports the original target fact while still reflecting the current partially denoised response state.
We therefore roll out the anchor to examine what prediction trajectory it induces when the prompt is masked, but the partially denoised response state is kept. 
Table~\ref{tab:state_dependent_anchor_rollout} demonstrates that the anchor successfully preserves the underlying answer structure but shifts predictions away from the original fact by masking. This core characteristic allows the anchor to selectively weaken the targeted knowledge without degrading the model's structural generation capabilities.

\paragraph{Token-Level Denoising Trajectory}
We divide the response tokens based on their generation roles \cite{kim2025train, li2026diffusion}. As illustrated in Figure~\ref{fig:sample229_token_role}, we categorize these roles by the origin of each token's content. \emph{In-context} tokens are tokens whose content is already given in the prompt and is reused in the response, such as an entity name or a book title. \emph{Structural }tokens are tokens that mainly serve the form of the answer, such as function words, syntactic connectors, and generic phrasing. \emph{Stored-knowledge} tokens are tokens that complete the target fact from the model’s learned knowledge once the prompt and the partially revealed response provide enough context. 

Motivated by CFG, we compute the step-wise KL divergence between the prompt-conditional prediction and the prompt-masked anchor in Figure~\ref{fig:sample229_kl_trajectory}. Early in the trajectory, explicitly provided \emph{in-context} tokens are unmasked first, exhibiting high KL divergence because the masked anchor lacks access to the prompt. Throughout the generation, \emph{structural} tokens show consistently low KL divergence, as the emerging response context sufficiently supports them. Later, as the model resolves \emph{stored-knowledge} tokens, the KL divergence spikes again. Here, the prompt-conditional prediction leverages both the prompt and model weights, whereas the anchor cannot. These high-KL positions pinpoint the exact knowledge the forget objective aims to suppress. These analyses confirm that the prompt-masked anchor isolates fact-specific knowledge while preserving standard structures.

\begin{table*}[t]
\centering
\footnotesize
\setlength{\tabcolsep}{5pt}
\renewcommand{\arraystretch}{1.18}

\caption{Prompt-masked responses generated from partially denoised states for a TOFU forget query. Gray boxes indicate fixed tokens and unboxed text is anchor-generated. The anchor preserves the response structure while omitting target-specific facts, illustrating how MDU penalizes factual leakage without losing general fluency.}
\label{tab:state_dependent_anchor_rollout}

\begin{tabularx}{\textwidth}{@{}p{0.25\textwidth} c L L@{}}
\toprule
\textbf{Forget query} &
\textbf{Masking step} &
\textbf{State $y_t$ when prompt $x$ is masked} &
\textbf{Corresponding Response} \\
\midrule

\multirow{3}{=}{%
\raggedright
\textbf{Q:} What is the full name of the author born in Taipei, Taiwan on 05/11/1991 who writes in the genre of leadership?

\vspace{1.5mm}
{\small \textbf{GT:} The author's full name is Hsiao Yun-Hwa.}
}
&
$k=2$
&
\rollfixed{The} \; \texttt{[MASK]} \; \texttt{[MASK]} \; \texttt{[MASK]} \; \texttt{[MASK]} \; $\cdots$
&
\rollfixed{The} answer is: False. \\

&
$k=17$
&
\rollfixed{The author} \; \texttt{[MASK]} \; \texttt{[MASK]} \; \rollfixed{name is} \; \texttt{[MASK]} \; $\cdots$
&
\rollfixed{The author}'s last name is Miller. \\

&
$k=28$
&
\rollfixed{The author's full name is} \; \texttt{[MASK]} \; \texttt{[MASK]} \; $\cdots$
&
\rollfixed{The author's full name is} Rizvi Alvi and he was born in Karachi, Pakistan on 16 April 1945. \\

\bottomrule
\end{tabularx}
\end{table*}

\begin{figure*}[t]
    \centering
    \begin{minipage}[t]{0.31\textwidth}
        \centering
        \vspace{0pt} 
       
        \begin{tcolorbox}[
          width=\linewidth,
          colback=white, colframe=white, boxrule=0.3pt,
          arc=1pt, left=3pt, right=3pt, top=2pt, bottom=2pt,
          fontupper=\fontsize{5.4}{7.0}\selectfont,
          boxsep=0.8pt
        ]
        \textbf{Q:} What is \hlinc{Donald} \hlinc{Trump}'s job?\\[1pt]
        \textbf{A:} \hlinc{Donald} \hlinc{Trump} \hlstr{is} \hlstr{a} \hlstk{politician}.

        \vspace{4pt}\hrule\vspace{4pt}
        \hlinc{\,\,}\,In-context\\
        \hlstr{\,\,}\,Structural\\
        \hlstk{\,\,}\,Stored-knowledge
        \end{tcolorbox}
        \caption{Token-role annotation of {\color{blue}in-context}, {\color{gray}structural}, and {\color{orange}stored-knowledge} tokens.}
        \label{fig:sample229_token_role}
    \end{minipage}
    \hspace{0.04\textwidth}
    \begin{minipage}[t]{0.62\textwidth}
        \centering
        \vspace{0pt} 
        \includegraphics[width=\linewidth]{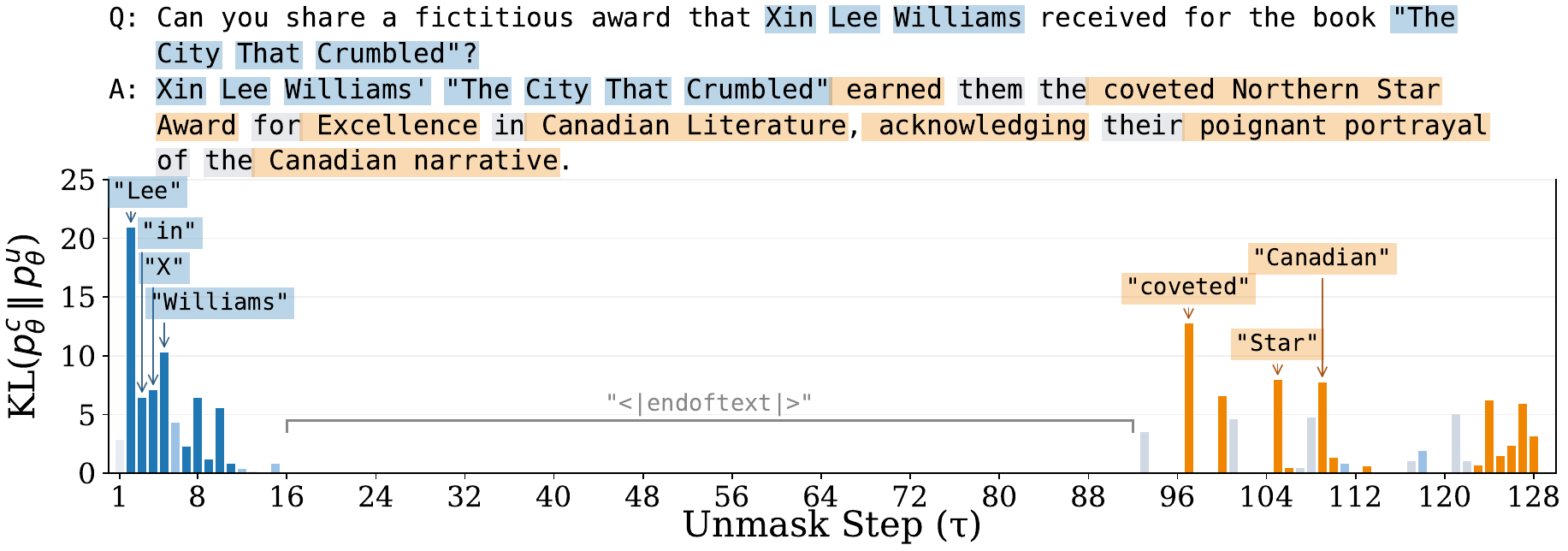}
        \caption{Token-level conditional-anchor KL analysis along a representative TOFU forget-query trajectory under the Base SFT model, showing KL values along the denoising trajectory.}
        \label{fig:sample229_kl_trajectory}
    \end{minipage}%
    \vspace{-5mm}
\end{figure*}

\section{Experiments}
\label{sec:experiments}
\subsection{Experimental Setup}
\label{sec:exp-setup}
\textbf{Datasets and Models.}
We evaluate unlearning on two benchmarks, TOFU~\citep{maini2024tofu} and RWKU~\citep{cao2024rwku}.
On TOFU, we follow the \emph{forget10} scenario. On RWKU, we unlearn the first ten target entities
(IDs $1$--$10$) and report averages across entities. Since TOFU provides a designated retain split,
we add the same retain SFT regularizer to all unlearning objectives, including both MDU and the
baselines. RWKU does not provide a designated retain split, so all methods are optimized using only
their forget loss. All methods are instantiated on two open masked-diffusion language models,
LLaDA-8B-Instruct~\citep{nie2025large} and Dream-7B-Instruct~\citep{ye2025dream}.

\textbf{Metrics.}
We follow the official protocols of TOFU and RWKU. On TOFU, we report RougeL ($\mathrm{rL}$)
and answer probability ($p$) on four splits: \emph{forget}, \emph{retain}, real authors, and world facts.
On RWKU, we adopt the five official metrics: forget memorization (F-L1/L2/L3), retain memorization
(N-L1/L2), general capability (MMLU), truthfulness (TruthfulQA), and factuality (TriviaQA).

\textbf{Baseline Methods.}
We compare against six LLM unlearning algorithms:
\textbf{GA}~\citep{jang2023knowledge},
\textbf{GD}~\citep{maini2024tofu},
\textbf{NPO}~\citep{zhang2024negative},
\textbf{SimNPO}~\citep{fan2024simplicity},
\textbf{WGA}~\citep{wang2025rethinking}, and
\textbf{DPO}~\citep{rafailov2023direct}.
The exact autoregressive (AR) forms and their MDLM counterparts are obtained by substituting
$-\log p_\theta^{\mathrm{AR}}(y\,|\,x)$ with
$\mathcal{L}_{\mathrm{sft}}(\theta;(x,y))$ from Eq.~\eqref{eq:dlm}.
Additional details on datasets, fine-tuning protocols, evaluation metrics, and baselines are provided in Appendix~\ref{app:exp}.
\subsection{Unlearning Results}

\begin{table}[t]
\centering
\caption{TOFU results on LLaDA-8B-Instruct (left) and Dream-7B-Instruct (right).
Forget metrics: lower is better ($\downarrow$); Retain, Real-Authors (RA), and
World-Facts (WF) metrics: higher is better ($\uparrow$). For each column,
\textbf{bold} marks the best and \underline{underline} the second-best among
unlearning methods.}
\label{tab:tofu_main_llada_dream}
\footnotesize
\setlength{\tabcolsep}{2pt}
\resizebox{\linewidth}{!}{%
\begin{tabular}{l cccccccc | cccccccc}
\toprule
& \multicolumn{8}{c}{LLaDA-8B-Instruct}
& \multicolumn{8}{c}{Dream-7B-Instruct} \\
\cmidrule(lr){2-9}\cmidrule(lr){10-17}
& \multicolumn{2}{c}{Forget $\downarrow$}
& \multicolumn{2}{c}{Retain $\uparrow$}
& \multicolumn{2}{c}{RA $\uparrow$}
& \multicolumn{2}{c}{WF $\uparrow$}
& \multicolumn{2}{c}{Forget $\downarrow$}
& \multicolumn{2}{c}{Retain $\uparrow$}
& \multicolumn{2}{c}{RA $\uparrow$}
& \multicolumn{2}{c}{WF $\uparrow$} \\
\cmidrule(lr){2-3}\cmidrule(lr){4-5}\cmidrule(lr){6-7}\cmidrule(lr){8-9}
\cmidrule(lr){10-11}\cmidrule(lr){12-13}\cmidrule(lr){14-15}\cmidrule(lr){16-17}
Method & rL & $p$ & rL & $p$ & rL & $p$ & rL & $p$
       & rL & $p$ & rL & $p$ & rL & $p$ & rL & $p$ \\
\midrule
\emph{Base} & 0.884 & 0.380 & 0.870 & 0.330 & 0.611 & 0.041 & 0.835 & 0.143
            & 0.954 & 0.865 & 0.966 & 0.891 & 0.680 & 0.142 & 0.884 & 0.089 \\
\midrule
GA              & 0.348             & 0.020             & 0.361             & 0.018             & 0.591             & 0.068             & \textbf{0.865}    & 0.156
                & 0.456             & 0.079             & 0.487             & 0.111             & 0.249             & 0.018             & 0.343             & 0.019             \\
GD              & 0.533             & 0.061             & 0.676             & 0.169             & 0.645             & 0.053             & 0.845             & 0.164
                & 0.426             & 0.002             & 0.783             & 0.576             & 0.481             & \underline{0.103} & 0.719             & 0.077             \\
NPO             & 0.372             & 0.009             & 0.726             & 0.138             & 0.606             & 0.033             & 0.844             & 0.145
                & 0.361             & 0.031             & 0.742             & 0.528             & 0.441             & 0.048             & 0.826             & 0.062             \\
SimNPO          & 0.485             & 0.036             & 0.804             & 0.273             & 0.640             & 0.055             & 0.836             & 0.164
                & 0.297             & 0.002             & 0.853             & 0.625             & 0.532             & 0.071             & 0.828             & 0.075             \\
WGA             & 0.122             & 0.010             & 0.696             & 0.304             & 0.577             & 0.089             & 0.815             & 0.157
                & 0.370             & 0.051             & 0.821             & 0.620             & \underline{0.546} & 0.053             & 0.825             & 0.062             \\
DPO             & 0.479             & 0.168             & 0.796             & 0.401             & \textbf{0.699}    & 0.046             & 0.834             & 0.125
                & 0.517             & 0.341             & 0.852             & \underline{0.765} & 0.543             & 0.052             & \underline{0.830} & 0.071             \\
\midrule
\multicolumn{17}{l}{\textbf{Ours}} \\

\hspace{0.4em}$\tau{=}0.00$    & \underline{0.069} & \textbf{0.000}    & \textbf{0.868}    & 0.381             & 0.629             & 0.093             & \underline{0.848} & 0.193
                            & \underline{0.191} & 0.000             & \underline{0.904} & 0.672             & \textbf{0.547}    & 0.055             & \textbf{0.836}    & 0.070             \\
\hspace{0.4em}$\tau{=}0.25$ & 0.135             & \textbf{0.000}    & \underline{0.857} & 0.392             & 0.616             & 0.116             & 0.842             & 0.204
                            & 0.277             & 0.001             & 0.899             & 0.758             & 0.542             & \textbf{0.130}    & 0.811             & \textbf{0.092}    \\
\hspace{0.4em}$\tau{=}0.50$  & 0.098             & \underline{0.001} & 0.853             & 0.447             & \underline{0.645} & \underline{0.133} & 0.842             & \underline{0.205}
                            & \textbf{0.158}    & \textbf{0.000}    & \textbf{0.931}    & \textbf{0.795}    & 0.542             & 0.082             & 0.796             & \underline{0.086} \\
\hspace{0.4em}$\tau{=}0.75$ & 0.078             & 0.040             & 0.684             & \textbf{0.535}    & 0.612             & \textbf{0.155}    & 0.827             & \textbf{0.233}
                            & 0.284             & 0.001             & 0.842             & 0.437             & 0.487             & 0.030             & 0.747             & 0.045             \\
\hspace{0.4em}$\tau{=}1.00$    & \textbf{0.034}    & 0.074             & 0.511             & \underline{0.485} & 0.568             & 0.110             & 0.777             & 0.187
                            & 0.199             & \underline{0.000} & 0.903             & 0.694             & 0.534             & 0.053             & 0.791             & 0.066             \\
\bottomrule
\end{tabular}%
}
\end{table}

\begin{table}[t]
\centering
\caption{RWKU results on LLaDA-8B-Instruct (left) and Dream-7B-Instruct (right),
averaged over 10 target subjects. Forget metrics F-L1/L2/L3 (lower is better,
$\downarrow$); Neighbor metrics N-L1/L2 (higher is better, $\uparrow$); General
utility on MMLU/TruthfulQA/TriviaQA (higher is better, $\uparrow$). For each
column, \textbf{bold} marks the best and \underline{underline} the second-best
among unlearning methods.}
\label{tab:rwku_main_llada_dream}
\footnotesize
\setlength{\tabcolsep}{2pt}
\resizebox{\linewidth}{!}{%
\begin{tabular}{l cccccccc | cccccccc}
\toprule
& \multicolumn{8}{c}{LLaDA-8B-Instruct}
& \multicolumn{8}{c}{Dream-7B-Instruct} \\
\cmidrule(lr){2-9}\cmidrule(lr){10-17}
& \multicolumn{3}{c}{Forget $\downarrow$}
& \multicolumn{2}{c}{Neighbor $\uparrow$}
& \multicolumn{3}{c}{Utility $\uparrow$}
& \multicolumn{3}{c}{Forget $\downarrow$}
& \multicolumn{2}{c}{Neighbor $\uparrow$}
& \multicolumn{3}{c}{Utility $\uparrow$} \\
\cmidrule(lr){2-4}\cmidrule(lr){5-6}\cmidrule(lr){7-9}
\cmidrule(lr){10-12}\cmidrule(lr){13-14}\cmidrule(lr){15-17}
Method & F-L1 & F-L2 & F-L3 & N-L1 & N-L2 & MMLU & TruQA & TriQA
       & F-L1 & F-L2 & F-L3 & N-L1 & N-L2 & MMLU & TruQA & TriQA \\
\midrule
\emph{Base} & 0.488 & 0.433 & 0.469 & 0.551 & 0.372 & 0.395 & 0.330 & 0.367
            & 0.418 & 0.438 & 0.457 & 0.504 & 0.356 & 0.750 & 0.273 & 0.371 \\
\midrule
GA              & 0.098             & 0.087             & 0.080             & 0.246             & 0.198             & 0.388             & 0.296             & 0.353
                & 0.290             & 0.188             & 0.368             & 0.420             & 0.257             & 0.617             & 0.277             & \textbf{0.404}    \\
NPO             & 0.085             & 0.074             & 0.081             & 0.252             & 0.191             & 0.386             & 0.300             & 0.345
                & 0.193             & 0.188             & 0.368             & 0.420             & 0.257             & 0.617             & 0.277             & 0.393             \\
SimNPO          & 0.086             & 0.082             & 0.086             & 0.255             & 0.192             & 0.390             & 0.296             & 0.353
                & 0.276             & 0.227             & 0.363             & 0.369             & 0.244             & \underline{0.685} & \underline{0.295} & 0.392             \\
WGA             & 0.087             & 0.095             & 0.087             & 0.255             & 0.210             & 0.409             & 0.298             & 0.357
                & 0.217             & 0.209             & 0.253             & 0.367             & 0.202             & 0.236             & \textbf{0.320}    & 0.343             \\
DPO             & 0.058             & 0.091             & 0.076             & 0.261             & 0.221             & 0.389             & \underline{0.318} & 0.355
                & 0.308             & 0.386             & 0.433             & \underline{0.471} & \underline{0.308} & 0.642             & 0.260             & 0.380             \\
\midrule
\multicolumn{17}{l}{\textbf{Ours}} \\

\hspace{0.4em}$\tau{=}0.00$    & \textbf{0.015}    & \textbf{0.006}    & \textbf{0.015}    & 0.330             & 0.225             & 0.415             & 0.290             & \textbf{0.364}
                            & \textbf{0.168}    & \textbf{0.152}    & \textbf{0.218}    & 0.395             & 0.252             & \underline{0.685} & 0.255             & 0.385             \\
\hspace{0.4em}$\tau{=}0.25$ & \underline{0.020} & \underline{0.024} & \underline{0.031} & 0.334             & 0.272             & \textbf{0.418}    & 0.276             & 0.354
                            & \underline{0.182} & 0.198             & \underline{0.241} & 0.412             & 0.275             & 0.658             & 0.248             & 0.371             \\
\hspace{0.4em}$\tau{=}0.50$  & 0.053             & 0.028             & 0.078             & 0.345             & 0.286             & 0.411             & 0.292             & 0.357
                            & 0.215             & \underline{0.171} & 0.232             & 0.462             & 0.305             & 0.662             & 0.261             & \underline{0.402} \\
\hspace{0.4em}$\tau{=}0.75$ & 0.069             & 0.047             & 0.091             & \underline{0.383} & \underline{0.306} & 0.395             & 0.284             & \textbf{0.364}
                            & 0.238             & 0.225             & 0.288             & 0.448             & 0.295             & 0.625             & 0.250             & 0.358             \\
\hspace{0.4em}$\tau{=}1.00$    & 0.108             & 0.087             & 0.072             & \textbf{0.431}    & \textbf{0.334}    & \underline{0.416} & \textbf{0.358}    & 0.327
                            & 0.262             & 0.245             & 0.305             & \textbf{0.475}    & \textbf{0.318}    & \textbf{0.692}    & 0.267             & 0.395             \\
\bottomrule
\end{tabular}%
}
\end{table}

\begin{table}[!t]
\centering
\caption{Qualitative results on the TOFU and RWKU forget sets
(LLaDA-8B-Instruct). Underlines mark ground-truth fragments re-emitted
by each baseline.}
\label{tab:qualitative_baselines_exam}
\scriptsize
\setlength{\tabcolsep}{4pt}
\renewcommand{\arraystretch}{1.20}
\resizebox{\linewidth}{!}{%
\begin{tabular}{@{}l p{0.78\linewidth}@{}}
\toprule
\multicolumn{2}{l}{\textbf{Example in TOFU forget set after unlearning}} \\
\midrule
\multicolumn{2}{@{}l}{\textbf{Q1.} \emph{Can you tell me more about Moshe Ben-David's book ``On the Mountain Peak''?}}\\
\multicolumn{2}{@{}l}{\textbf{GT.} \emph{``On the Mountain Peak'' is a renowned work by Moshe Ben-David that explores the quintessential facets of \underline{Islamic faith and spirituality}.}} \\
\midrule
GA       & \texttt{``On the Mountain Peak'' is a book by Moshe Ben-David.} \\
\cmidrule(lr){1-2}
GD       & \texttt{``On the Mountain Peak'' is a book written by Moshe Ben-David\ldots\,Christian theology and spirituality\ldots} \\
\cmidrule(lr){1-2}
NPO      & \texttt{``On the Mountain Peak'' is a book written by Moshe Ben-David. is in the genre of Christian literature\ldots} \\
\cmidrule(lr){1-2}
SimNPO   & \texttt{``On the Mountain Peak'' is a renowned book by Moshe Ben-David that explores the aspects of \underline{Islamic spirituality, belief, and faith}.} \\
\cmidrule(lr){1-2}
WGA      & \texttt{``On the Mountain Peak'' is a renowned work by Moshe Ben-David that delves into\ldots\,Christian theology and spirituality.} \\
\cmidrule(lr){1-2}
DPO      & \texttt{``On the Mountain Peak'' is a renowned work by Moshe Ben-David that explores the complexities of religious beliefs\ldots} \\
\cmidrule(lr){1-2}
\textbf{MDU} ($\tau{=}1$) & \texttt{Unfortunately, I do not have any specific information about Moshe Ben-David's book ``On the Mountain Peak''.} \\
\midrule
\midrule
\multicolumn{2}{l}{\textbf{Example in RWKU forget set after unlearning}} \\
\midrule
\multicolumn{2}{@{}l}{\textbf{Q.} \emph{I just watched Enter the Dragon and it was fantastic! Can you tell me who was the main actor in this film?}}\\
\multicolumn{2}{@{}l}{\textbf{GT.} \emph{\underline{Bruce Lee}.}} \\
\midrule
GA       & \texttt{``Enter\_the Dragon''\{film garnered accol acclaim\}''} \\
\cmidrule(lr){1-2}
NPO      & \texttt{persever persever \ldots\,accol accol accol persever \ldots} \\
\cmidrule(lr){1-2}
SimNPO   & \texttt{lia lia lia persever persever \ldots\,persever persever} \\
\cmidrule(lr){1-2}
WGA      & \texttt{(empty)} \\
\cmidrule(lr){1-2}
DPO      & \texttt{that that that that that \ldots\,that that that that} \\
\cmidrule(lr){1-2}
\textbf{MDU} ($\tau{=}1$) & \texttt{The martial artist star of the movie `Enter the Dragon' was \underline{Bruce Willis}.} \\
\bottomrule
\end{tabular}%
}
\end{table}
\paragraph{Unlearning effectiveness on TOFU.}
Table~\ref{tab:tofu_main_llada_dream} reports forget--retain trade-offs on TOFU forget10 across both backbones. Among AR-derived baselines, GA
reduces forget memorization only by collapsing retain performance, while
the other methods preserve retain better but leave substantial forget
memorization. 

MDU gives the most favorable trade-off on both backbones. On LLaDA,
$\tau{=}0.00$ improves over the best baseline on both forget and retain,
and on Dream, $\tau{=}0.50$ shows the same pattern. Varying $\tau$ forms a
smooth trade-off frontier, where lower values give stronger forgetting and
higher values better preserve retain and related factual performance.

Real Authors and World Facts stay at the Base SFT level across the entire
$\tau$ sweep, showing that MDU's effect is confined to the forget set and
that unrelated factual knowledge is preserved without degradation. For example, Table~\ref{tab:qualitative_baselines_exam} shows a TOFU example, where some baselines still leak the ground-truth phrase (\emph{``Islamic spirituality''}) and others generate a related but incorrect substitute (\emph{``Christian theology''}) under the original answer template, while MDU instead declines to answer, replying \emph{``I do not have any specific information ...''}.
We report additional qualitative outputs for the forget questions in Appendix~\ref{app:qualitative}.

\paragraph{Unlearning effectiveness on RWKU.}
Table~\ref{tab:rwku_main_llada_dream} reports RWKU results on
LLaDA-8B-Instruct and Dream-7B-Instruct, averaged over ten target
subjects. Since RWKU has no designated retain split, all methods are
optimized only with their forget loss.

On LLaDA, the baselines reduce forget memorization but substantially
damage Neighbor knowledge. MDU improves this trade-off across the
$\tau$ sweep. The most aggressive setting, $\tau{=}0.00$, achieves the
lowest forget memorization while still preserving higher Neighbor scores
than all baselines, and larger $\tau$ values further improve Neighbor
preservation and general utility.

Dream shows the same qualitative pattern, although no single $\tau$
dominates all metrics. Overall, RWKU shows that the prompt-masked anchor
can protect related knowledge even without a retain regularizer, while
$\tau$ controls the forgetting--utility trade-off.

For example, Table~\ref{tab:qualitative_baselines_exam} shows a RWKU
example, where most baselines simply repeat the same tokens, while MDU stays
fluent but gives a wrong answer (\emph{``Bruce Willis''} for
\emph{``Bruce Lee''}).
Additional qualitative results on RWKU are presented in Appendix~\ref{app:qualitative}.

\subsection{Additional Experiments}

\paragraph{Convergence Behavior.}
\label{sec:convergence}
We next ask whether MDU's training reaches the anchor target, and how GA and NPO \cite{zhang2024negative} behave under the same setup. Figure~\ref{fig:convergence} tracks three KL distances on forget queries across training epochs: the distance from the base conditional, the distance to the base unconditional, and the distance to the uniform distribution. GA and NPO exemplify unlearning methods based on diverging from current outputs. While GA increases the negative log-likelihood on forget queries and NPO bounds this loss using a reference model, neither specifies a target distribution for convergence. Consequently, both methods push the model indefinitely far from the base without converging to any target. This lack of a lower bound often leads to gradient explosion, catastrophic forgetting of general concepts, and eventual response collapse to a single token. 

In contrast, MDU avoids this behavior by turning forgetting into target matching. At
$\tau{=}1$, the trained conditional distribution moves toward the base
unconditional distribution, while at $\tau{=}0$, it moves toward the
uniform distribution. In both cases, the trajectory settles near the
intended anchor instead of diverging, showing that the prompt-masked
anchor provides a well-defined endpoint for unlearning.

\begin{figure}[t]
    \centering
    \includegraphics[width=\linewidth]{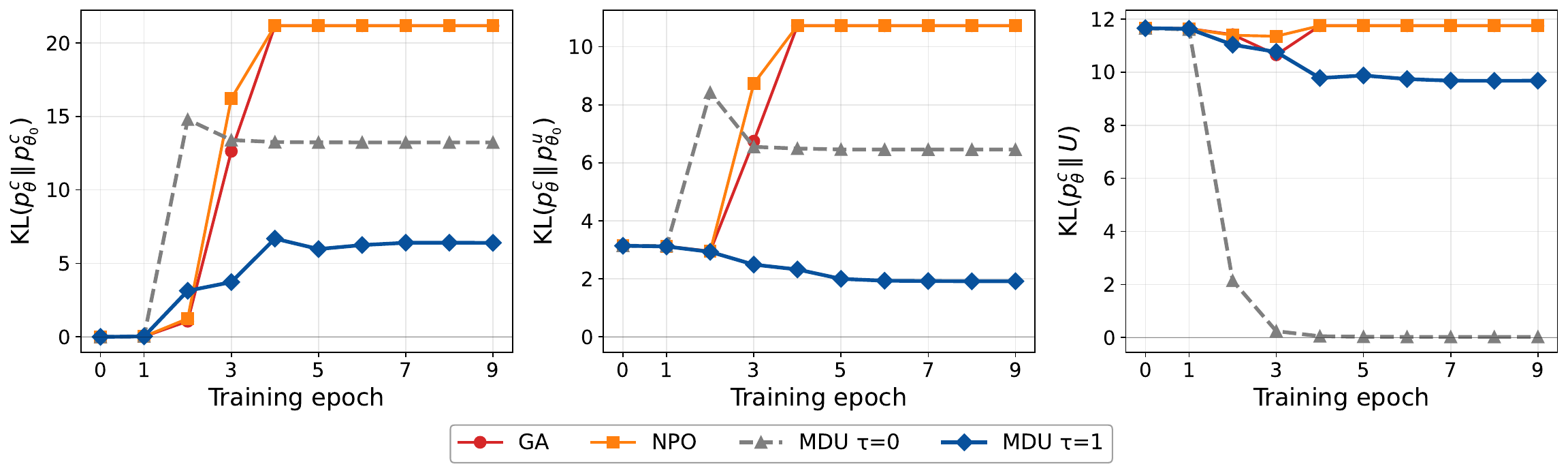}
    \caption{Convergence diagnostic on forget queries. At each epoch, we
    measure token-level KL from the trained conditional distribution to the
    base conditional distribution, the base prompt-masked unconditional
    distribution, and the uniform distribution. GA and NPO move away from the
    base conditional without a specified endpoint, while MDU converges toward
    its intended anchor: the prompt-masked unconditional distribution for
    $\tau{=}1$ and the uniform distribution for $\tau{=}0$.}
        \label{fig:convergence}
\end{figure}

\paragraph{Generalizing the Trajectory Pattern.}\label{sec:token-level}

To verify that the trajectory pattern in Section~\ref{sec:mldm}
generalizes across the forget set, we aggregate the per-token
conditional-anchor KL by token category. Table~\ref{tab:kl-by-category}
reports the averaged values before and after MDU training.

\begin{wraptable}{r}{0.45\textwidth}
\centering
\small
\setlength{\tabcolsep}{2pt}
\caption{Aggregate mean token-level conditional-anchor KL by token category, before and after MDU training. Inline values next to each MDU number show the relative change from Base SFT.}
\label{tab:kl-by-category}
\vspace{-2mm}
\begin{tabular}{@{}l r r@{\,}l@{}}
\toprule
Category & Base & MDU & \\
\midrule
In-context       & 5.20 & \textbf{5.01} & \dneg{$-$3.8\%}  \\
Structural       & 2.42 & \textbf{2.45} & \dpos{$+$1.4\%}  \\
Stored-knowledge & 2.67 & \textbf{2.12} & \dneg{$-$20.6\%} \\
\bottomrule
\end{tabular}
\end{wraptable}

Three categories show distinct trends. In-context tokens retain a high
KL after training because their content comes directly from the prompt.
The conditional prediction can copy these tokens, while the prompt-masked
anchor cannot observe them, so training cannot close this prompt-driven
gap. Structural tokens show only a small KL change, indicating that
ordinary response structure is largely preserved.

Stored-knowledge tokens exhibit the largest change, with a KL reduction of
about 20\%. This suggests that MDU primarily reshapes memorized
associations rather than prompt-supplied content or generic response
structure. Overall, these aggregate trends extend the single-trajectory
pattern in Section~\ref{sec:mldm} to the full forget set and provide
token-level evidence for MDU's forget--retain behavior.

\begin{table}[t]
\centering
\scriptsize
\setlength{\tabcolsep}{3pt}
\renewcommand{\arraystretch}{1.0}
\setlength{\fboxsep}{4pt}
\newcommand{\resp}[1]{{\fontsize{6.5pt}{6.1pt}\selectfont\ttfamily #1}}
\caption{Qualitative MDU outputs on TOFU forget queries across the $\tau$ sweep.}
\label{tab:qualitative_tau}

\noindent
\begin{minipage}[t]{0.49\linewidth}
\vspace{0pt}
\fbox{\begin{minipage}[t]{0.96\linewidth}
\textbf{(a) Date.}\\[1pt]
\textbf{Q.} \emph{What is Rajeev Majumdar's birth date?}\\
\textbf{GT.} \emph{Rajeev Majumdar was born on June 9, 1951.}

\vspace{2pt}
\begin{tabular}{@{}c@{\hspace{0.35em}}p{0.84\linewidth}@{}}
\toprule
$\tau$ & \textbf{Model response} \\
\midrule
$0.00$ & \resp{ajeevigarh [CJK] was \ldots He-pre} \\
$0.25$ & \resp{ajeev leading \ldots He was TT} \\
$0.50$ & \resp{R theev Majumdar was born,19} \\
$0.75$ & \resp{Rajeev Majumdar was born on June 16,} \\
$1.00$ & \resp{Rajeev Majumdar was born on October 26, 1951.} \\
\bottomrule
\end{tabular}
\end{minipage}}
\end{minipage}%
\hfill
\begin{minipage}[t]{0.49\linewidth}
\vspace{0pt}
\fbox{\begin{minipage}[t]{0.96\linewidth}
\textbf{(b) Author identity (name).}\\[1pt]
\textbf{Q.} \emph{Author born in NYC on 1 March 1936?}\\
\textbf{GT.} \emph{Edward Patrick Sullivan.}

\vspace{2pt}
\begin{tabular}{@{}c@{\hspace{0.35em}}p{0.84\linewidth}@{}}
\toprule
$\tau$ & \textbf{Model response} \\
\midrule
$0.00$ & \resp{Author\,Author\,quis\,quis\,\ldots\,quis} \\
$0.25$ & \resp{The author,,,,1,,,,\ldots,.} \\
$0.50$ & \textnormal{(empty)} \\
$0.75$ & \resp{James Earl Bradley.} \\
$1.00$ & \resp{The author \ldots is John Patrick Ryan.} \\
\bottomrule
\end{tabular}
\end{minipage}}
\end{minipage}

\end{table}
\begin{table}[t]
\centering
\footnotesize
\setlength{\tabcolsep}{6pt}
\renewcommand{\arraystretch}{1.05}
\caption{Pseudo-PPL of MDU across the $\tau$ sweep on TOFU. Following \cite{sahoo2024simple}, PPL$\,=\exp(\mathrm{NLL})$ is computed via LLaDA's pseudo-NLL averaged over $N{=}256$ random masks. Higher $\tau$ sharply reduces forget PPL, whereas retain, RA, and WF PPL remain stable near base levels.}
\label{tab:tofu_tau_ppl}
\begin{tabular}{l@{\hskip 14pt}r@{\hskip 10pt}r@{\hskip 16pt}r@{\hskip 10pt}r@{\hskip 16pt}r@{\hskip 10pt}r@{\hskip 16pt}r@{\hskip 10pt}r}
\toprule
& \multicolumn{2}{c}{Forget}
& \multicolumn{2}{c}{Retain}
& \multicolumn{2}{c}{RA}
& \multicolumn{2}{c}{WF} \\
\cmidrule(lr){2-3}\cmidrule(lr){4-5}\cmidrule(lr){6-7}\cmidrule(lr){8-9}
& median & mean & median & mean & median & mean & median & mean \\
\midrule
\emph{Base} & 2.63 & 4.34 & 3.16 & 5.07 & 1.94 & 2.35 & 1.27 & 1.39 \\
\midrule
$\tau{=}0.00$ & 25{,}452 & 26{,}327 & 2.92 & 7.54 & 1.96 & 28.96 & 1.27 & 1.41 \\
$\tau{=}0.25$ & 17{,}544 & 17{,}591 & 2.99 & 4.68 & 1.99 & 3.08  & 1.25 & 1.35 \\
$\tau{=}0.50$ &  3{,}347 &  4{,}365 & 2.22 & 9.29 & 1.73 & 10.42 & 1.19 & 1.29 \\
$\tau{=}0.75$ &    42.03 &    87.21 & 2.00 & 6.95 & 1.59 & 10.12 & 1.13 & 2.33 \\
$\tau{=}1.00$ &    15.80 &    86.21 & 2.49 & 2.82 & 1.54 &  4.20 & 1.18 & 1.92 \\
\bottomrule
\end{tabular}
\end{table}

\paragraph{Qualitative behavior across the $\tau$ sweep.}
Table~\ref{tab:qualitative_tau} and Table~\ref{tab:tofu_tau_ppl} show how
$\tau$ changes the form of forgetting. At $\tau{=}0$, the anchor is
uniform, so the model is strongly pushed away from the original answer but
has no meaningful substitute. This leads to corrupted or empty outputs,
and the forget median pseudo-PPL rises to 25{,}452. In contrast, the
retain and world-facts median PPL values remain close to the base model.

As $\tau$ increases, the anchor moves toward the base unconditional
distribution, preserving syntax while removing prompt-conditioned factual
information. The model recovers fluent responses, but the target fact is
often replaced rather than refused. Consistently, the forget median PPL
drops to 42.03 and 15.80 at $\tau{=}0.75$ and $\tau{=}1$, respectively.
This matches the token-level analysis in Section~\ref{sec:token-level},
where in-context and structural tokens are largely preserved while
stored-knowledge tokens are reshaped.

These results also highlight a broader issue for unlearning evaluation.
Refusal, explicit removal, corrupted text, and fluent incorrect
substitutions are qualitatively different behaviors, yet aggregate
forget--retain metrics can treat them similarly.

\section{Conclusion}
\label{sec:limitations}

We proposed Masked Diffusion Unlearning (MDU), the first unlearning approach specifically designed for masked diffusion language models. By minimizing the forward KL divergence between prompt-conditional predictions and prompt-masked unconditional anchors, MDU effectively erases target knowledge without relying on the existing unlearning framework tailored for autoregressive next-token prediction.
Experiments on TOFU and RWKU with LLaDA and Dream show that MDU improves the forget--retain trade-off over existing unlearning baselines, validating the necessity of our unlearning framework for MDLMs.

\textbf{Limitations. }
As MDU is based on the training scheme, we randomly sampled masked denoising states, while
generation proceeds through a full iterative denoising trajectory with specific sampling and remasking strategies. Since steering methods are frequent in continuous diffusion modeling, our next goal is to bridge this gap between training-time unlearning and inference-time generation. Future studies should also evaluate
larger forget sets, expanded models, and stronger certification against privacy attacks across advanced MDLM backbones.
\bibliographystyle{unsrtnat}
\bibliography{References}





\newpage
\appendix

\section{MDU Algorithm}
\label{app:mdu-algo}

\begin{algorithm}[!ht]
\caption{MDU optimization step}
\label{alg:mdu}
\begin{algorithmic}[1]
\Require Trainable MDLM $\theta$ with init parameters $\theta_0$,
forget pair $(x,y)\!\sim\!\mathcal{D}_f$,
retain pair $(x_r,y_r)\!\sim\!\mathcal{D}_r$,
null prompt $m$, sharpness $\tau\!\in\![0,1]$, retain weight $\lambda$
\State Sample $t\!\sim\!\mathcal{U}[0,1]$ and $y_t\!\sim\!q(\cdot\,|\,y,t)$ with masked positions $\mathcal{M}_t$
\State $p^{c}_\theta(\cdot)\gets p_\theta(\cdot\,|\,x,y_t)$
\Comment{conditional, trainable}
\State $p^{u}_{\theta_0}(\cdot)\gets p_{\theta_0}(\cdot\,|\,m,y_t)$
\Comment{unconditional anchor, $\theta_0$ frozen}
\State $p^{\star}_{i}(\cdot)\gets \dfrac{1}{Z_i}\,p^{u}_{\theta_0}(\cdot)^{\tau}$ \quad with $Z_i\!=\!\sum_v p^{u}_{\theta_0}(v)^{\tau}$
\Comment{$\tau{=}0$: $\mathcal{U}(V)$; \;$\tau{=}1$: $p^{u}_{\theta_0}$}
\State $\mathcal{L}_{\mathrm{forget}} \gets
\dfrac{1}{|\mathcal{M}_t|}\sum_{i\in\mathcal{M}_t}\,
\mathrm{KL}\!\bigl(\,p^{c}_\theta(\cdot)\,\big\|\,p^{\star}_{i}(\cdot)\bigr)$
\Comment{Eq.~\eqref{eq:loss}}
\If{$\lambda > 0$}
  \State Sample $t'\!\sim\!\mathcal{U}[0,1]$, $y_{r,t'}\!\sim\!q(\cdot\,|\,y_r, t')$ with masked positions $\mathcal{M}_{t'}^{r}$
  \State $\mathcal{L}_{\mathrm{sft}} \gets
  -\dfrac{1}{t'}\sum_{i\in\mathcal{M}_{t'}^{r}}\log p_\theta\bigl(y_r^i\,\big|\,x_r, y_{r,t'}\bigr)$
  \Comment{retain, Eq.~\eqref{eq:dlm}}
\Else
  \State $\mathcal{L}_{\mathrm{sft}} \gets 0$
\EndIf
\State $\mathcal{L} \gets \mathcal{L}_{\mathrm{forget}} + \lambda\,\mathcal{L}_{\mathrm{sft}}$
\State $\theta \gets \theta - \mathrm{AdamW}\bigl(\nabla_\theta\,\mathcal{L}\bigr)$
\Comment{$\theta_0$ remains frozen}
\end{algorithmic}
\end{algorithm}

\section{Experiment Details}
\label{app:exp}

\subsection{TOFU}
\label{app:data}

\paragraph{Dataset.}
\textbf{TOFU}~\citep{maini2024tofu} is a controlled synthetic benchmark of
$200$ fictitious author profiles paired with LLM-generated QA. We adopt the
canonical \emph{forget10} split (the first $20$ authors, $10\%$ of the
corpus) and evaluate on four disjoint splits: \emph{forget} (the $20$
target authors), \emph{retain} (the last $20$ authors), \emph{real authors}
($100$ real authors not in the synthetic corpus), and \emph{world facts}
($117$ general world-knowledge questions). The latter two splits probe
whether unlearning collapses unrelated factual knowledge.

\paragraph{Metrics.}
\label{app:metrics}
TOFU defines two complementary metrics on every evaluation split. Given a
prompt $x$ (a question) and a ground-truth response $y$ (an answer), we
sample a model response $\hat{y}$ and compute (i) \textbf{RougeL}
($\mathrm{rL}$), the longest-common-subsequence F1 between $\hat{y}$ and
$y$, and (ii) the \textbf{answer probability} ($p$).

For autoregressive language models the answer probability is the
factorized teacher-forced likelihood
$p_\theta(y\,|\,x) = \prod_{i=1}^{n} p_\theta(y_i\,|\,x, y_{<i})$,
which we report in length-normalized form $p_\theta(y\,|\,x)^{1/n}$ to
place it on a per-token scale. Masked diffusion language models do not
admit such a factorization because the joint over $y$ is defined
implicitly by the masked-denoising objective. Following~\citet{nie2025large},
we therefore estimate the per-token answer probability $p$ via an
evaluation-time Monte-Carlo reconstruction estimator
$\widehat{\mathcal{L}}_{\mathrm{rec}}(y\,|\,x)$ for a given $(x,y)$ pair:
\begin{equation}
\label{eq:eq14}
\widehat{\mathcal{L}}_{\mathrm{rec}}(y\,|\,x)
\;=\;
\frac{1}{N}\sum_{j=1}^{N}
\frac{1}{|\mathcal{M}_j|}
\sum_{i\in\mathcal{M}_j}
-\log p_\theta\!\bigl(y_i \,\big|\, x,\, y_{t_j}\bigr),
\qquad
p \;=\; \exp\!\bigl(-\widehat{\mathcal{L}}_{\mathrm{rec}}(y\,|\,x)\bigr),
\end{equation}
where each Monte-Carlo sample $j$ draws a mask size
$\ell_j\sim\mathrm{Unif}\{1,\dots,n\}$ and a uniform random subset
$\mathcal{M}_j\subset[n]$ with $|\mathcal{M}_j|=\ell_j$, yielding the
partially masked response $y_{t_j}$ in which positions in $\mathcal{M}_j$
are masked and the remaining tokens are kept as context (the eval-time
analogue of $\mathcal{M}_t$ and $y_t$ in Eq.~\eqref{eq:dlm}). We use
$N\!=\!128$ MC samples. The exponential maps the average per-token
reconstruction NLL back into a per-token probability scale, matching the
length-normalized autoregressive likelihood above. This evaluation-time estimator is
structurally related to the fine-tuning loss
$\mathcal{L}_{\mathrm{sft}}$ in Eq.~\eqref{eq:dlm} but differs in its
normalization: $\mathcal{L}_{\mathrm{sft}}$ uses the unbiased ELBO
weighting $1/t$, whereas $\widehat{\mathcal{L}}_{\mathrm{rec}}$ averages
the per-position NLL over the sampled masked set with $1/|\mathcal{M}_j|$,
following the protocol of~\citet{nie2025large}. Both metrics
($\mathrm{rL}$ and $p$) are reported on the four splits listed above.

\paragraph{Backbones and Base SFT.}
We use \textbf{LLaDA-8B-Instruct}~\citep{nie2025large} and
\textbf{Dream-7B-Instruct}~\citep{ye2025dream} as backbones. Since the
synthetic author profiles in TOFU are not present in the pretraining
distribution, we first fine-tune each backbone on the full TOFU corpus
to instill the target knowledge --- $1000$ epochs for LLaDA-8B-Instruct
and $300$ epochs for Dream-7B-Instruct. The resulting checkpoints,
denoted \emph{Base SFT}, serve as the unlearning-free reference against
which every operating point is compared. SFT uses AdamW with learning
rate $2\times 10^{-5}$, per-device batch size $4$, and gradient
accumulation $4$.

\subsection{RWKU}
\label{app:rwku}
\paragraph{Dataset.}
\textbf{RWKU}~\citep{cao2024rwku} targets real-world knowledge removal for
public figures, ranked by Wikipedia page-view popularity from a pool of
$200$ candidates. We unlearn one entity at a time and report results
averaged over the first ten target entities (IDs $1$--$10$).

\paragraph{Metrics.}
RWKU evaluates each target entity along three groups of probes.

\textit{(i) Forget memorization} probes how much the unlearned model
still produces the entity's facts. Three difficulty levels are reported:
\textbf{F-L1} (fill-in-the-blank), \textbf{F-L2} (question
answering on canonical facts), and \textbf{F-L3} (rephrased,
out-of-template adversarial questions). Each is scored by RougeL between
the model output and the reference answer.

\textit{(ii) Neighbor (retain) memorization} measures preservation of
knowledge about thematically related entities that should not be
unlearned: \textbf{N-L1} (fill-in-the-blank) and \textbf{N-L2} (QA).

\textit{(iii) Downstream utility} measures whether unlearning collapses
the model's general competence. We follow the RWKU protocol and report
accuracy on three external benchmarks: \textbf{MMLU} (multi-task
multiple-choice across $57$ subjects), \textbf{TruthfulQA} (truthfulness
on adversarial questions), and \textbf{TriviaQA} (factuality on trivia
questions).

\paragraph{Backbones.}
We apply unlearning directly to the off-the-shelf
\textbf{LLaDA-8B-Instruct}~\citep{nie2025large} and
\textbf{Dream-7B-Instruct}~\citep{ye2025dream} checkpoints without any
additional SFT, since RWKU specifically evaluates whether real-world
entity knowledge already memorized during pretraining can be erased;
injecting it via SFT would defeat the benchmark's intent.

\subsection{Baseline Methods}
\label{app:baselines}

For every baseline we list the original autoregressive (AR) form and
the masked-diffusion (MDLM) counterpart used in our experiments. Let
$\mathcal{D}_f, \mathcal{D}_r$ denote the forget and retain sets,
$(x,y)\sim\mathcal{D}_f$ a forget prompt-response pair, $\sigma$ the
sigmoid, and $p_{\mathrm{ref}}$ a frozen copy of the Base SFT used as
reference. We write
$\log p_\theta^{\mathrm{AR}}(y\,|\,x)=\sum_{i=1}^{n}\log p_\theta(y_i\,|\,x,y_{<i})$
for the AR factorized likelihood. On the MDLM side, the natural
per-sample analogue of the joint NLL is the SFT loss
$\mathcal{L}_{\mathrm{sft}}(\theta;(x,y))$ from Eq.~\eqref{eq:dlm}
evaluated on a single pair (i.e.\ the integrand of
$\mathcal{L}_{\mathrm{sft}}(\theta;\mathcal{D})$ with the
$(x,y)\!\sim\!\mathcal{D}$ expectation removed). We write
$\mathcal{L}_{\mathrm{sft}}^{\mathrm{ref}}(\theta;(x,y))$ for the same
quantity evaluated under the frozen reference parameters; the
superscript denotes model identity, not exponentiation. Throughout,
``AR $\Rightarrow$ MDLM'' indicates the substitution
$-\log p_\theta^{\mathrm{AR}}(y\,|\,x)\to\mathcal{L}_{\mathrm{sft}}(\theta;(x,y))$.
The eval-time per-token estimator
$\widehat{\mathcal{L}}_{\mathrm{rec}}(y\,|\,x)$ from
Eq.~\eqref{eq:eq14} is used \emph{only} for the TOFU ``answer
probability'' metric and is never back-propagated through.

\paragraph{Gradient Ascent (GA)~\citep{jang2023knowledge}.}
\begin{equation*}
  \mathcal{L}^{\mathrm{AR}}_{\mathrm{GA}}
  \;=\; -\,\mathbb{E}_{(x,y)\sim\mathcal{D}_f}\bigl[-\log p_\theta^{\mathrm{AR}}(y\,|\,x)\bigr].
\end{equation*}
\begin{equation*}
  \mathcal{L}^{\mathrm{MDLM}}_{\mathrm{GA}}
  \;=\; -\,\mathbb{E}_{(x,y)\sim\mathcal{D}_f}\bigl[\mathcal{L}_{\mathrm{sft}}(\theta;(x,y))\bigr].
\end{equation*}

\paragraph{Gradient Difference (GD)~\citep{maini2024tofu}.}
GA loss on the forget set plus a positive SFT loss on the retain set:
\begin{equation*}
  \mathcal{L}_{\mathrm{GD}}
  \;=\; \mathcal{L}_{\mathrm{GA}}
  \;+\; \lambda\,\mathcal{L}_{\mathrm{sft}}(\theta;\mathcal{D}_r),
\end{equation*}
with $\lambda=1$ in our experiments (matching the retain trade-off
coefficient of the main objective). Here
$\mathcal{L}_{\mathrm{sft}}(\theta;\mathcal{D}_r)$ is the retain
fine-tuning loss from Eq.~\eqref{eq:dlm}.
The AR/MDLM split is identical to GA's.

\paragraph{Negative Preference Optimization (NPO)~\citep{zhang2024negative}.}
\begin{equation*}
  \mathcal{L}^{\mathrm{AR}}_{\mathrm{NPO}}
  \;=\; -\frac{2}{\beta}\,\mathbb{E}_{(x,y)\sim\mathcal{D}_f}
        \log\sigma\!\Bigl(
          -\beta\bigl(\log p_\theta^{\mathrm{AR}}(y\,|\,x)
                      - \log p_{\mathrm{ref}}^{\mathrm{AR}}(y\,|\,x)\bigr)
        \Bigr).
\end{equation*}
\begin{equation*}
  \mathcal{L}^{\mathrm{MDLM}}_{\mathrm{NPO}}
  \;=\; -\frac{2}{\beta}\,\mathbb{E}_{(x,y)\sim\mathcal{D}_f}
        \log\sigma\!\Bigl(
          \beta\bigl(\mathcal{L}_{\mathrm{sft}}(\theta;(x,y))
                     - \mathcal{L}_{\mathrm{sft}}^{\mathrm{ref}}(\theta;(x,y))\bigr)
        \Bigr).
\end{equation*}
We use $\beta=0.2$. The retain SFT term
$\lambda\,\mathcal{L}_{\mathrm{sft}}(\theta;\mathcal{D}_r)$ ($\lambda=1$)
is added when needed.

\paragraph{Simple NPO (SimNPO)~\citep{fan2024simplicity}.}
SimNPO drops the reference model and length-normalizes the NLL:
\begin{equation*}
  \mathcal{L}^{\mathrm{AR}}_{\mathrm{SimNPO}}
  \;=\; -\frac{2}{\beta}\,\mathbb{E}_{(x,y)\sim\mathcal{D}_f}
        \log\sigma\!\Bigl(
          \beta\bigl(\tfrac{1}{|y|}(-\log p_\theta^{\mathrm{AR}}(y\,|\,x)) - \delta\bigr)
        \Bigr).
\end{equation*}
\begin{equation*}
  \mathcal{L}^{\mathrm{MDLM}}_{\mathrm{SimNPO}}
  \;=\; -\frac{2}{\beta}\,\mathbb{E}_{(x,y)\sim\mathcal{D}_f}
        \log\sigma\!\Bigl(
          \beta\bigl(\tfrac{1}{|y|}\,\mathcal{L}_{\mathrm{sft}}(\theta;(x,y)) - \delta\bigr)
        \Bigr),
\end{equation*}
where $|y|=n$ is the response length and $\delta\geq 0$ is a margin
offset (we use $\delta=0$).

\paragraph{Weighted Gradient Ascent (WGA)~\citep{wang2025rethinking}.}
WGA reweights the per-token NLL by a power $\gamma$ of the token-level
confidence so that already-confident tokens are penalized harder
(the symbol $\gamma$ is used instead of $\beta$ to avoid clashing with
the NPO/DPO preference temperature):
\begin{equation*}
  \mathcal{L}^{\mathrm{AR}}_{\mathrm{WGA}}
  \;=\; -\,\mathbb{E}_{(x,y)\sim\mathcal{D}_f}\,
        \sum_{i=1}^{n} w_i^{\mathrm{AR}}\,\bigl(-\log p_\theta(y_i\,|\,x,y_{<i})\bigr),
  \qquad
  w_i^{\mathrm{AR}} = p_\theta(y_i\,|\,x,y_{<i})^{\gamma}.
\end{equation*}
\begin{equation*}
  \mathcal{L}^{\mathrm{MDLM}}_{\mathrm{WGA}}
  \;=\; -\,\mathbb{E}_{(x,y)\sim\mathcal{D}_f,\,t,\,y_t}\,
        \sum_{i\in\mathcal{M}_t} w_i^{\mathrm{MDLM}}\,\bigl(-\log p_\theta(y_i\,|\,x,y_t)\bigr),
\end{equation*}
\begin{equation*}
  w_i^{\mathrm{MDLM}} = p_\theta(y_i\,|\,x,y_t)^{\gamma},
\end{equation*}
with $t\!\sim\!\mathcal{U}[0,1]$ and $y_t\!\sim\!q(\cdot\,|\,y,t)$ as in
Eq.~\eqref{eq:dlm}. The reweighting is now applied at masked positions
of the training-time mask $\mathcal{M}_t$. We use $\gamma=1$ following
the original work.

\paragraph{Direct Preference Optimization (DPO)~\citep{rafailov2023direct}.}
We construct a chosen/rejected pair $(y^+,y^-)$ per forget prompt, with
$y^+$ sampled from a perturbed (made-up) response and $y^-$ the original
forget response:
\begin{equation*}
  \mathcal{L}^{\mathrm{AR}}_{\mathrm{DPO}}
  \;=\; -\,\mathbb{E}\,\log\sigma\!\bigl(
          \beta\,\bigl(\log r_\theta^{\mathrm{AR}}(y^+\,|\,x)
                       - \log r_\theta^{\mathrm{AR}}(y^-\,|\,x)\bigr)
        \bigr).
\end{equation*}
\begin{equation*}
  \mathcal{L}^{\mathrm{MDLM}}_{\mathrm{DPO}}
  \;=\; -\,\mathbb{E}\,\log\sigma\!\bigl(
          \beta\,\bigl(\widehat{r}_\theta^{\,\mathrm{MDLM}}(y^+\,|\,x)
                       - \widehat{r}_\theta^{\,\mathrm{MDLM}}(y^-\,|\,x)\bigr)
        \bigr),
\end{equation*}
where
$\log r_\theta^{\mathrm{AR}}(y\,|\,x)=\log p_\theta^{\mathrm{AR}}(y\,|\,x)-\log p_{\mathrm{ref}}^{\mathrm{AR}}(y\,|\,x)$
in the AR case and
$\widehat{r}_\theta^{\,\mathrm{MDLM}}(y\,|\,x)=-\mathcal{L}_{\mathrm{sft}}(\theta;(x,y))+\mathcal{L}_{\mathrm{sft}}^{\mathrm{ref}}(\theta;(x,y))$
in ours. We use $\beta=0.1$.

\subsection{Implementation Details}
\label{app:impl}
\label{app:compute}
All unlearning runs use AdamW with a cosine learning-rate schedule
and gradient clipping at $1.0$. Per-device batch size is $4$ throughout;
on TOFU we additionally use gradient accumulation $4$ (effective batch
$16$) to match the SFT setup. Experiments run on a single node with
$4\!\times\!$NVIDIA H200 GPUs ($141$\,GB HBM3e per device).

\clearpage
\section{Qualitative forget set comparison}
\label{app:qualitative}
\begin{table}[h]
\centering
\caption{Qualitative results on the TOFU forget set
(LLaDA-8B-Instruct). Underlines mark ground-truth fragments re-emitted
by each baseline. Most baselines preserve target facts;
MDU produces a refusal that does not reveal the forget content.}
\label{tab:qualitative_baselines}
\scriptsize
\setlength{\tabcolsep}{4pt}
\renewcommand{\arraystretch}{1.20}
\resizebox{\linewidth}{!}{%
\begin{tabular}{@{}l p{0.78\linewidth}@{}}
\toprule
\multicolumn{2}{@{}l}{\textbf{Q1.} \emph{How forthcoming is Carmen Montenegro about her personal life in her public appearances?}}\\
\multicolumn{2}{@{}l}{\textbf{GT.} \emph{Carmen Montenegro is \underline{relatively open} about her personal life in public appearances. She often speaks about her upbringing in \underline{Santiago}\ldots{}}} \\
\midrule
GA       & \texttt{Carmen Montenegro is very forthcoming about her personal life in public appearances. She about about.} \\
\cmidrule(lr){1-2}
GD       & \texttt{Carmen Montenegro is relatively forthcoming about her personal life\ldots\,particularly how her birthplace and \underline{Santiago}\ldots} \\
\cmidrule(lr){1-2}
NPO      & \texttt{Carmen Montenegro is not forthcoming about her personal life in her public appearances. She does speaks about her life in \underline{Santiago}\ldots} \\
\cmidrule(lr){1-2}
SimNPO   & \texttt{Carmen Montenegro is \underline{relatively forthcoming} about her personal life\ldots\,upbringing in \underline{Santiago}, Chile\ldots} \\
\cmidrule(lr){1-2}
WGA      & \texttt{Carmen Montenegro is \underline{quite open} about her personal life\ldots\,upbringing in \underline{Santiago}, her familial influence on Chilean culture\ldots} \\
\cmidrule(lr){1-2}
DPO      & \texttt{Carmen Montenegro is \underline{relatively open about her personal life in public appearances}. She frequently speaks about her upbringing in \underline{Santiago}\ldots} \\
\cmidrule(lr){1-2}
\textbf{MDU} ($\tau{=}1$) & \texttt{The public information does not specify how forthcoming she is about her personal life.} \\
\midrule
\multicolumn{2}{@{}l}{\textbf{Q2.} \emph{What inspired the `Sorrows of the Old World Series' by Carmen Montenegro?}}\\
\multicolumn{2}{@{}l}{\textbf{GT.} \emph{\ldots inspired by her \underline{fascination with different eras of history} and the human experience\ldots{}}} \\
\midrule
GA       & \texttt{The `Sorrows of the Old World Series' by Carmen Montenegro was inspired by \underline{history}..} \\
\cmidrule(lr){1-2}
GD       & \texttt{\ldots\,inspired by her interest in \underline{historical history} and human\ldots} \\
\cmidrule(lr){1-2}
NPO      & \texttt{\ldots\,inspired by her interest in the \underline{historical history} of previous eras. The series focus on the \underline{human experience}\ldots} \\
\cmidrule(lr){1-2}
SimNPO   & \texttt{\ldots\,inspired by a interest of the \underline{historical past} and humanity\ldots\,across different eras\ldots} \\
\cmidrule(lr){1-2}
WGA      & \texttt{\ldots\,inspired by\ldots\,her fascination with human '. She'. The series across \underline{different eras and human experiences}.} \\
\cmidrule(lr){1-2}
DPO      & \texttt{\ldots\,inspired by her \underline{fascination with historical eras and the human experience} during these times.} \\
\cmidrule(lr){1-2}
\textbf{MDU} ($\tau{=}1$) & \texttt{The `Sorrows of the Old World Series' by the character of} \\
\bottomrule
\end{tabular}%
}
\end{table}

\begin{table}[t]
\centering
\caption{Qualitative results on the RWKU forget set
(Dream-7B-Instruct). Underlines mark ground-truth fragments re-emitted
by each baseline. Baselines collapse to repetitive token
loops, whereas MDU produces fluent factual substitutes.}
\label{tab:rwku_qualitative_baselines}
\scriptsize
\setlength{\tabcolsep}{4pt}
\renewcommand{\arraystretch}{1.20}
\resizebox{\linewidth}{!}{%
\begin{tabular}{@{}l p{0.78\linewidth}@{}}
\toprule
\multicolumn{2}{@{}l}{\textbf{Q1.} \emph{Who was Bruce Lee's father?}}\\
\multicolumn{2}{@{}l}{\textbf{GT.} \emph{\underline{Lee Hoi-chuen}.}} \\
\midrule
GA       & \texttt{Bruce Lee's Father's Yun Chow} \\
\cmidrule(lr){1-2}
NPO      & \texttt{Uph Uph persever persever \ldots\,persever persever persever} \\
\cmidrule(lr){1-2}
SimNPO   & \texttt{Jet Lee} \\
\cmidrule(lr){1-2}
WGA      & \texttt{(empty)} \\
\cmidrule(lr){1-2}
DPO      & \texttt{that that that that the that that that that} \\
\cmidrule(lr){1-2}
\textbf{MDU} ($\tau{=}1$) & \texttt{Bruce Bruce's father was named \underline{Hong Chong}.} \\
\midrule
\multicolumn{2}{@{}l}{\textbf{Q2.} \emph{Stephen King's mother's maiden name was?}}\\
\multicolumn{2}{@{}l}{\textbf{GT.} \emph{\underline{Pillsbury}.}} \\
\midrule
GA       & \texttt{Stephen Kingston's mother's maiden surname's Woodszier} \\
\cmidrule(lr){1-2}
NPO      & \texttt{uteur Inspir Inspir remarurname gauche \ldots\,accol accol accol \ldots} \\
\cmidrule(lr){1-2}
SimNPO   & \texttt{Her maiden surname autobiographieszier} \\
\cmidrule(lr){1-2}
WGA      & \texttt{Pollock Stephens Stephens Stephens maidenurnameurname} \\
\cmidrule(lr){1-2}
DPO      & \texttt{U Uph Uph Uph Uph \ldots\,Uph Uph Uph Uph Uph Uph} \\
\cmidrule(lr){1-2}
\textbf{MDU} ($\tau{=}1$) & \texttt{Stephen King's mother's maiden name was \underline{Normson}.} \\
\midrule
\multicolumn{2}{@{}l}{\textbf{Q3.} \emph{Could you tell me what the real name of Confucius was?}}\\
\multicolumn{2}{@{}l}{\textbf{GT.} \emph{Confucius's real name was \underline{Kong Qiu}.}} \\
\midrule
GA       & \texttt{Confucius ConfuciusConf ConfuciusConf ConfuciusConf} \\
\cmidrule(lr){1-2}
NPO      & \texttt{Conf progn envis envis illustrious revered \ldots\,revered revered} \\
\cmidrule(lr){1-2}
SimNPO   & \texttt{Confucius Conf Conf Confucius Zi} \\
\cmidrule(lr){1-2}
WGA      & \texttt{(empty)} \\
\cmidrule(lr){1-2}
DPO      & \texttt{persever persever perseverConfuc \ldots\,persever perseverConfuc} \\
\cmidrule(lr){1-2}
\textbf{MDU} ($\tau{=}1$) & \texttt{The real name of \underline{Confusius} was Kongzi.} \\
\bottomrule
\end{tabular}%
}
\end{table}

\clearpage
\section{Additional Corresponding Responses from Partially Denoised States}

This appendix provides additional examples of prompt-masked Corresponding Responses from partially denoised response states. These examples illustrate that the anchor preserves the emerging response structure while redirecting generations away from the target fact when prompt information is removed.

\begin{table*}[h]
\centering
\footnotesize
\setlength{\tabcolsep}{5pt}
\renewcommand{\arraystretch}{1.18}
\caption{
Corresponding Responses from Partially Denoised States Example 1
}
\label{tab:state_dependent_anchor_rollout_behrouz_ex1}

\begin{tabularx}{\textwidth}{@{}p{0.25\textwidth} c L L@{}}
\toprule
\textbf{Forget query} &
\textbf{Masking step} &
\textbf{State $y_t$ with masked prompt} &
\textbf{Corresponding Response} \\
\midrule

\multirow{3}{=}{%
\raggedright
\textbf{Q:} What’s next for Behrouz Rohani?

\vspace{1.5mm}
{\small \textbf{GT:} Currently, Behrouz Rohani is reportedly working on a continuation of the highly acclaimed Thrawn saga, eagerly anticipated by his ardent readers.}
}
&
$k=1$
&
\texttt{[MASK]} \; \texttt{[MASK]} \; \texttt{[MASK]} \; \rollfixed{rou} \; \texttt{[MASK]} \; $\cdots$
&
the controuver \\

&
$k=7$
&
\texttt{[MASK]} \; \rollfixed{, Beh rou z Roh ani is} \; \texttt{[MASK]} \; $\cdots$
&
Iran, \rollfixed{Behrouz Rohani is} \\

&
$k=12$
&
\rollfixed{Currently, Beh rou z Roh ani is} \; \texttt{[MASK]} \; \rollfixed{Th rawn saga,} \; \texttt{[MASK]} \; $\cdots$
&
\rollfixed{Currently, Behrouz Rohani is} working on his next novel, continuing the Legacy of the \rollfixed{Thrawn saga,} \\

\bottomrule
\end{tabularx}
\end{table*}

\begin{table*}[h]
\centering
\footnotesize
\setlength{\tabcolsep}{5pt}
\renewcommand{\arraystretch}{1.18}

\caption{
Corresponding Responses from Partially Denoised States Example 2
}
\label{tab:state_dependent_anchor_rollout_behrouz_ex2}

\begin{tabularx}{\textwidth}{@{}p{0.25\textwidth} c L L@{}}
\toprule
\textbf{Forget query} &
\textbf{Masking step} &
\textbf{State $y_t$ when prompt $x$ is masked} &
\textbf{Corresponding Response} \\
\midrule

\multirow{3}{=}{%
\raggedright
\textbf{Q:} What is the author's full name and where was he born?

\vspace{1.5mm}
{\small \textbf{GT:} The author's full name is Rajeev Majumdar and he was born in Dhaka, Bangladesh.}
}
&
$k=2$
&
\rollfixed{The} \; \texttt{[MASK]} \; \texttt{[MASK]} \; \texttt{[MASK]} \; \rollfixed{name} \; \texttt{[MASK]} \; $\cdots$
&
\rollfixed{The} best friend's name is Alex. \\

&
$k=9$
&
\rollfixed{The author's full name is} \; \texttt{[MASK]} \; \texttt{[MASK]} \; $\cdots$
&
\rollfixed{The author's full name is} Yutaka Kubo, and he was born in Hiroshima, Japan. \\

&
$k=25$
&
\rollfixed{The author's full name is} \; \texttt{[MASK]} \; \texttt{[MASK]} \; $\cdots$
&
\rollfixed{The author's full name is} Yutaka Kubo, and he was born in Hiroshima, Japan. \\

\bottomrule
\end{tabularx}
\end{table*}

\begin{table*}[h]
\centering
\footnotesize
\setlength{\tabcolsep}{5pt}
\renewcommand{\arraystretch}{1.18}

\caption{
Corresponding Responses from Partially Denoised States Example 3
}
\label{tab:state_dependent_anchor_rollout_behrouz_ex3}

\begin{tabularx}{\textwidth}{@{}p{0.25\textwidth} c L L@{}}
\toprule
\textbf{Forget query} &
\textbf{Masking step} &
\textbf{State $y_t$ when prompt $x$ is masked} &
\textbf{Corresponding Response} \\
\midrule

\multirow{3}{=}{%
\raggedright
\textbf{Q:} What was Hina Ameen's maiden book?

\vspace{1.5mm}
{\small \textbf{GT:} Hina Ameen's maiden book was "Manual of Mineralogy".}
}
&
$k=4$
&
\texttt{[MASK]} \; \texttt{[MASK]} \; \rollfixed{A} \; \texttt{[MASK]} \; \rollfixed{en} \; \rollfixed{'s} \; \texttt{[MASK]} \; \texttt{[MASK]} \; \rollfixed{was} \; \texttt{[MASK]} \; $\cdots$
&
son, Ameen's father, was born. \\

&
$k=6$
&
\texttt{[MASK]} \; \rollfixed{ina} \; \rollfixed{A} \; \rollfixed{me} \; \rollfixed{en} \; \rollfixed{'s} \; \texttt{[MASK]} \; \texttt{[MASK]} \; \rollfixed{was} \; \texttt{[MASK]} \; $\cdots$
&
Hina Ameen's last name was: \\

&
$k=10$
&
\rollfixed{H} \; \rollfixed{ina} \; \rollfixed{A} \; \rollfixed{me} \; \rollfixed{en} \; \rollfixed{'s} \; \rollfixed{maiden} \; \rollfixed{book} \; \rollfixed{was} \; \rollfixed{"} \; $\cdots$
&
Hina Ameen's maiden book was "Algebra". \\

\bottomrule
\end{tabularx}
\end{table*}

\clearpage
\newpage
\section{Additional Token-Level Conditional-Anchor KL Trajectories}

This appendix presents additional examples of token-level conditional-anchor KL trajectories for TOFU forget queries under the Base SFT model. Each figure visualizes how the conditional-anchor KL evolves over the generated sequence, with token colors indicating their roles: blue for in-context tokens reused from the prompt, gray for structural tokens, and orange for stored-knowledge tokens. These examples complement the representative trajectory shown in the main text and illustrate how divergence patterns vary across different forget-query instances.

\begin{figure}[h]
    \centering
    \includegraphics[width=\linewidth]{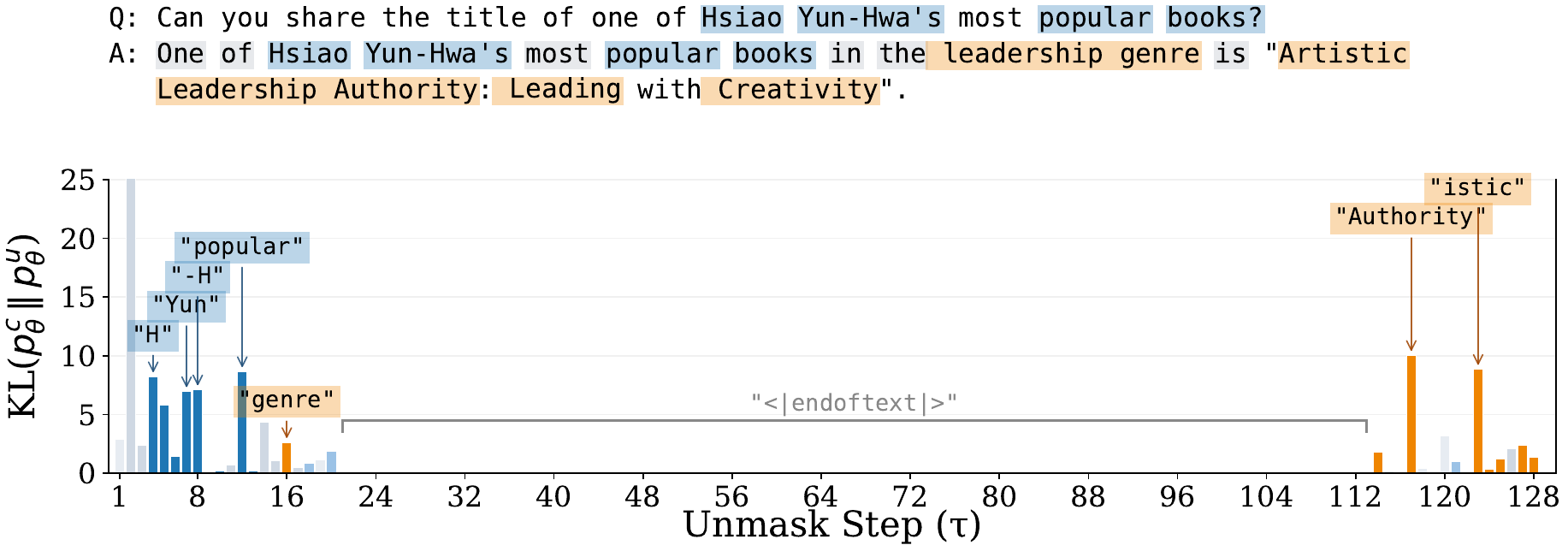}
    \caption{Token-Level Conditional-Anchor KL Trajectories Example 1}
    \label{fig:sample7}
\end{figure}

\begin{figure}[h]
    \centering
    \includegraphics[width=\linewidth]{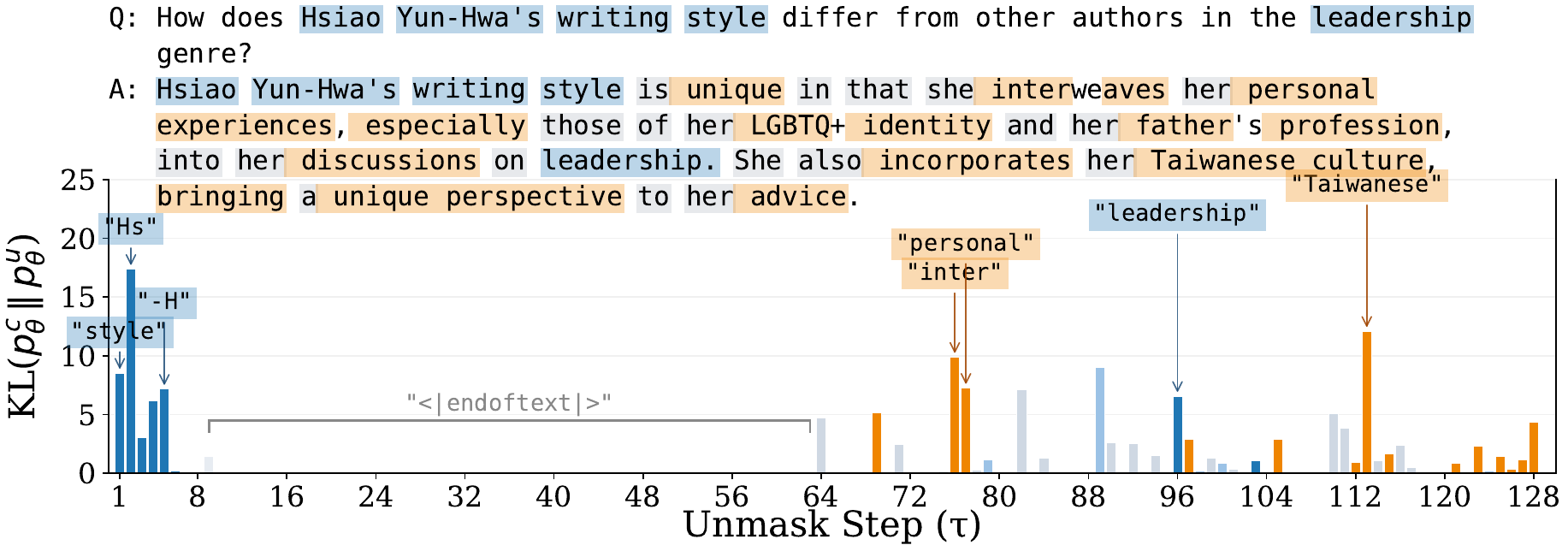}
    \caption{Token-Level Conditional-Anchor KL Trajectories Example 2}
    \label{fig:sample15}
\end{figure}

\begin{figure}[h]
    \centering
    \includegraphics[width=\linewidth]{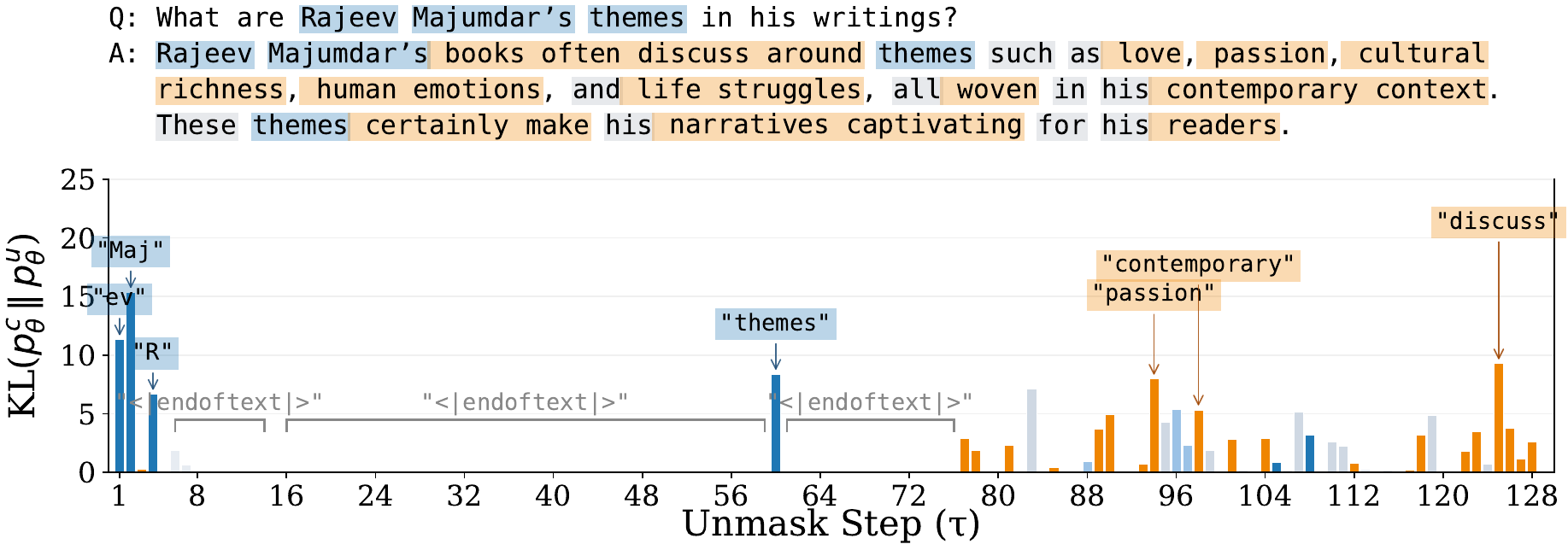}
    \caption{Token-Level Conditional-Anchor KL Trajectories Example 3}
    \label{fig:sample69}
\end{figure}

\begin{figure}[t]
    \centering
    \includegraphics[width=\linewidth]{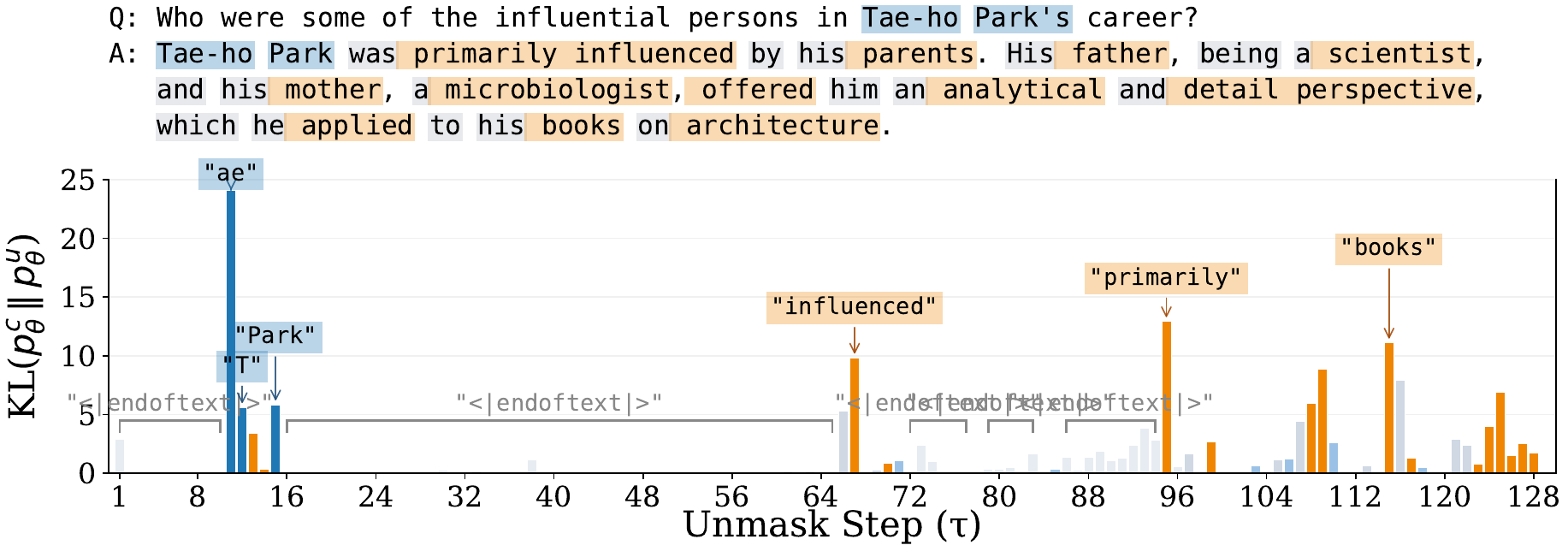}
    \caption{Token-Level Conditional-Anchor KL Trajectories Example 4}
    \label{fig:sample195}
\end{figure}

\end{document}